\documentclass{article}

\usepackage{booktabs,multirow}
\usepackage{caption}
\usepackage{placeins}
\usepackage{textcomp}

\usepackage{adjustbox}
\usepackage{acl}
\usepackage{dsfont}
\usepackage{listings}
\usepackage{amsmath,amssymb,amsfonts}
\usepackage{graphicx}
\graphicspath{{figures/}}
\usepackage{subcaption}
\usepackage{natbib}
\usepackage[most]{tcolorbox}
\newcommand{\OURS}{PivotAttack}
\title{PivotAttack: Rethinking the Search Trajectory in Hard-Label Text Attacks via Pivot Words}

\author{
Yuzhi Liang\thanks{\ \ Equal contribution.}\textsuperscript{,}\thanks{\ \ Corresponding author.}, 
  Shiliang Xiao\footnotemark[1],
  Jingsong Wei, 
  Qiliang Lin \and 
  Xia Li \\
  Guangdong University of Foreign Studies, China \\
  \texttt{yzliang@gdufs.edu.cn, silan2499776755@163.com, 1056111087@qq.com} \\
  \texttt{2043664242@qq.com, xiali@gdufs.edu.cn }
}

\usepackage{algorithm}
\usepackage[noend]{algpseudocode}
\usepackage{float}
\usepackage{enumitem}

\floatstyle{ruled}
\restylefloat{algorithm}

\begin{document}

\maketitle
\begin{abstract}
Existing hard-label text attacks often rely on inefficient "outside-in" strategies that traverse vast search spaces. We propose {\OURS}, a query-efficient "inside-out" framework. It employs a Multi-Armed Bandit  algorithm to identify \textit{Pivot Sets}—combinatorial token groups acting as prediction anchors—and strategically perturbs them to induce label flips. This approach captures inter-word dependencies and minimizes query costs. Extensive experiments across traditional models and Large Language Models demonstrate that {\OURS} consistently outperforms state-of-the-art baselines in both Attack Success Rate and query efficiency. 


\end{abstract}

\section{Introduction}

Deep neural networks have achieved remarkable performance across natural language processing tasks, yet they remain highly vulnerable to adversarial examples—small, often imperceptible perturbations that induce misclassification. Among various threat models, the hard-label black-box setting represents the most restrictive and realistic scenario: the attacker queries the target model and receives only a discrete class label, without access to gradients, confidence scores, or internal states. This constraint poses a significant challenge: \textit{how to generate semantically faithful adversarial examples with minimal queries?}

Existing hard-label attacks typically rely on approximating the decision boundary, but they suffer from intrinsic inefficiencies. \textbf{First}, many state-of-the-art methods, such as HyGloadAttack~\citep{DBLP:journals/nn/LiuXLYLZX24} and TextHoaxer~\citep{ye2022texthoaxer}, adopt an "outside-in" initialization strategy. They often start with a heavily perturbed text far from the original semantics and iteratively refine it to approach the decision boundary. This trajectory traverses a vast search space, inevitably consuming excessive queries and degrading textual quality.
\textbf{Second}, methods like VIWHard~\citep{DBLP:journals/ijon/ZhangWGZWL25} and LimeAttack~\citep{DBLP:conf/aaai/ZhuZ0WL24} attempt to identify important words via local surrogates but typically score tokens independently. This independence assumption ignores the combinatorial nature of language, often highlighting functional words while missing multi-word semantic anchors, leading to suboptimal perturbation sets. \textbf{Finally}, most methods lack interpretability, relying on opaque continuous relaxations or complex heuristic searches that offer little insight into why specific substitutions trigger a label flip.

To address these challenges, we propose {\OURS}\footnote{Our code is anonymously available at \href{https://anonymous.4open.science/r/PivotAttack-A20231003121/}{https://anonymous.4open.science/r/PivotAttack-A20231003121/}}, a query-efficient attack that fundamentally shifts the paradigm from "approximating the boundary" to "breaking the load-bearing walls." Specifically, {\OURS} implements a novel "inside-out" strategy. It starts from the original text and identifies a \textit{Pivot Set}—a compact group of tokens that anchors the model’s prediction. We observe that as long as this Pivot Set remains intact, the prediction is robust; however, strategically perturbing these tokens triggers a disproportionate collapse in model confidence, efficiently driving the instance across the decision boundary.

Methodologically, we formulate Pivot Set identification as a Multi-Armed Bandit (MAB) problem, employing the KL-LUCB algorithm to rigorously estimate the influence of token combinations under a limited budget. This rigorous formulation allows {\OURS} to distinguish true semantic anchors from statistical noise.

Our contributions are summarized as follows:
\begin{itemize}
    \item We propose the novel "inside-out" strategy, which attacks pivot words to advance toward the decision boundary from within the label-invariant region. This approach is significantly more query-efficient than mainstream "outside-in" methods that require expensive refinement steps.
    \item Unlike methods that rank tokens in isolation, {\OURS} explicitly accounts for inter-word interactions when selecting perturbations, enabling the identification of effective multi-word edits.
    \item We formulate Pivot Set selection via a multi-armed bandit framework, which generates human-readable intermediate outputs at each iteration, thereby improving both the interpretability and traceability of the attack behavior.
\end{itemize}

Extensive experiments verify that {\OURS} consistently outperforms baselines across varying architectures. Notably, on Large Language Models, {\OURS} demonstrates exceptional efficacy: it exposes the high vulnerability of zero-shot models and, more importantly, remains the most effective attacker against robust Fine-tuned LLMs, surpassing state-of-the-art methods in both success rate and query efficiency.

\section{Related Work}

Research on textual adversarial attacks is formally categorized based on the adversary's knowledge of the victim model.

\paragraph{White-box and Soft-label Attacks.}
White-box attacks assume full transparency, allowing direct optimization via gradients. \citet{DBLP:conf/acl/EbrahimiRLD18} proposed HotFlip for gradient-based character perturbations, while \citet{DBLP:conf/emnlp/GuoSJK21} introduced GBDA to optimize adversarial distributions via Gumbel-Softmax. More recently, TextGrad \citep{DBLP:conf/iclr/HouJZZ00C23} utilized gradients for precise robustness assessment.
In the black-box setting, soft-label attacks rely on output probabilities. Early approaches focused on synonym replacement strategies guided by lexical resources like HowNet and WordNet \citep{ren-etal-2019-generating, DBLP:conf/acl/ZangQYLZLS20}. Others optimize word selection through importance ranking (TextFooler; \citealp{JinJZS20}), Bayesian search \citep{DBLP:journals/corr/abs-2206-08575}, or conditional generative models \citep{DBLP:conf/emnlp/LiSLKWZHL23}. Recently, \citet{DBLP:conf/acl/0002XB25} proposed ALGEN, leveraging cross-model alignment for few-shot embedding inversion.

\paragraph{Hard-label Black-box Attacks.}
This setting is the most challenging, as attackers can access only the final discrete prediction. Existing methodologies generally adopt evolutionary or boundary-approximation strategies. 
Population-based methods, such as HLBB \citep{DBLP:conf/aaai/MaheshwaryMP21}, utilize genetic algorithms to evolve candidates but are often query-intensive. 
Boundary-approximation methods aim to locate and traverse the decision boundary. LeapAttack \citep{DBLP:conf/kdd/YeCMWM22} exploits directional cues, while GeoAttack \citep{DBLP:conf/coling/MengW20} and TextHoaxer \citep{ye2022texthoaxer} optimize within geometric or continuous embedding spaces. To mitigate local optima, HyGloadAttack \citep{DBLP:journals/nn/LiuXLYLZX24} introduces a hybrid optimization framework with perturbation matrices.
Refinement-based methods often start with effective substitutions or noise. TextHacker \citep{YuWC022} combines hybrid local search with attack history, and LimeAttack \citep{DBLP:conf/aaai/ZhuZ0WL24} employs local surrogate models (LIME) to estimate token importance. Most recently, VIWHard \citep{DBLP:journals/ijon/ZhangWGZWL25} utilized masked language models to identify critical words.

\section{Problem Formulation}
\label{sec:problem_formulation}

Given a victim classifier $f : \mathcal{X} \rightarrow \mathcal{Y}$ and an input $X$, our goal is to generate an adversarial example $X'$ that misleads the model (i.e., $f(X') \neq f(X)$) while preserving the semantic content of the original input. In the hard-label black-box setting, the attacker queries the victim model to obtain only the predicted label, without access to gradients or confidence scores. Accordingly, the objective is to find such a semantically consistent adversarial example $X'$ within a limited query budget $B$.


\section{Methodology}




Before detailing the algorithm, we clarify the intuition behind {\OURS}. Unlike traditional text attack methods that focus on identifying tokens to flip the predicted label immediately, {\OURS} targets robustness anchors—a specific set of tokens whose preservation ensures label stability. Conceptually, these tokens function as the "load-bearing walls" of the prediction: even if the majority of the sentence remains semantically intact, perturbing the Pivot Set often triggers a disproportionate collapse in model confidence. 

\begin{figure*}[!tbp]
  \centering
  \hspace*{-0.05\linewidth}
  \begin{subfigure}{\linewidth}
    \includegraphics[width=\linewidth]{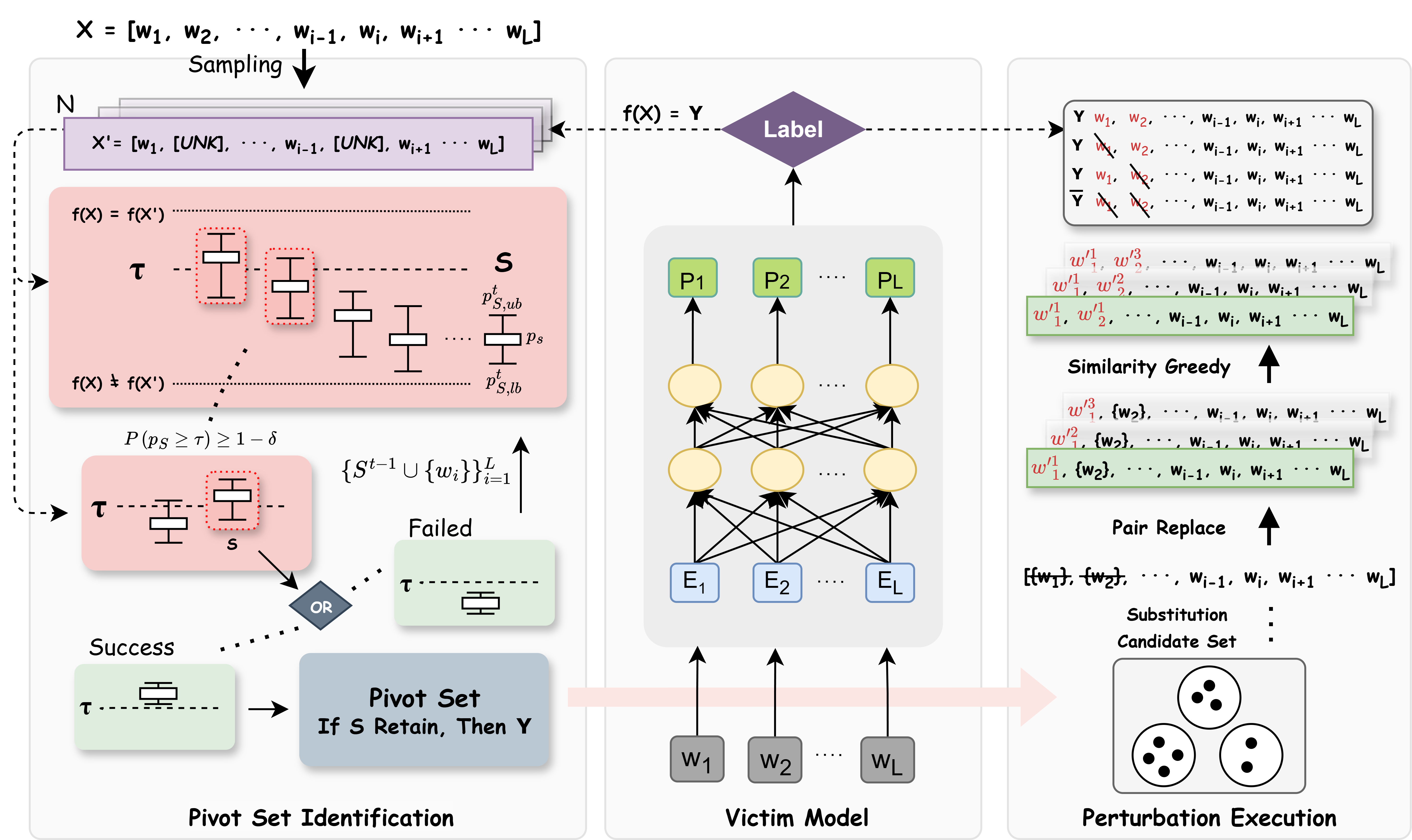}
  \end{subfigure}\hfill
  \caption{\textbf{The overall workflow of {\OURS}.} The framework first identifies Pivot Sets that anchor the model’s prediction, selecting sets with high retention precision and refining them via a multi-armed bandit. It then generates substitutions for the pivot words and selects the variant most similar to the original sentence as the final adversarial example.}
  \label{pic_overview}
\end{figure*}

Specifically, {\OURS} operates in two stages: (1) Pivot Set Identification, where a multi-armed bandit strategy is employed to isolate pivot words that anchor the model’s prediction (implying that perturbing non-pivot words leaves the output largely unchanged); and (2) Perturbation Execution, where synonym substitutions are applied specifically to these identified pivot words to generate adversarial samples. The overall workflow is illustrated in Figure~\ref{pic_overview}.

\subsection{Pivot Set Identification} 


The goal of this stage is to identify a Pivot Set $S = \{w_1, w_2, \dots, w_n\} $ such that, for a given input $ X $, when all words in $ S $ remain unperturbed in a new sample $ X' $, the model’s prediction is very likely to remain unchanged (i.e., $ f(X) = f(X') $). Formally, for a given input $x$ and a candidate set $S$, we estimate the retention precision $p_S$, which quantifies the probability that the victim model $f$ maintains its original prediction when non-pivot words are perturbed. This is defined as:
\begin{equation}
\label{eq:p_s}
p_{S} = \mathbb{E}_{\mathcal{D}(z \mid S)} \left[ \mathds{1}\left(f(x) = f(z)\right) \right]
\end{equation}
\noindent where $\mathds{1}(\cdot)$ is the indicator function, and $ z $ denotes samples drawn from $ \mathcal{D}(\,z \mid S) $, the distribution of perturbed texts that preserve all pivot words in $ S $. The final Pivot Set is required to satisfy $ p_{S} > \tau $, where $ \tau $ is a predefined threshold. A larger $ p_{S} $ indicates that preserving $ S $ makes it more likely for the predicted label to remain unchanged, implying that $ S $ is a stronger candidate for the Pivot Set.

It is generally intractable to compute $p_{S}$ exactly. To address this, we allow an approximation with a tolerable error $\delta \in [0,1]$ and adopt a probabilistic definition. Specifically, we regard $S$ as a Pivot  Set if $p_{S}$ satisfies the following condition:
\begin{equation}
    P\left( p_{S} \geq \tau \right) \geq 1 - \delta
\end{equation}
where $\delta$ is a predefined parameter controlling the acceptance threshold.

Multiple Pivot Sets may satisfy the criterion. In such cases, we prioritize those with fewer constituent words. This preference arises because manipulating fewer tokens yields adversarial examples that closely resemble the original sentence, enhancing their stealthiness. In summary, the search for a Pivot Set can be formulated as the following optimization problem:
\begin{equation}
\min_{S \: s.t. \: P\left( p_{S} \geq \tau \right) \geq 1 - \delta} |S|
\label{eq:obj}
\end{equation}




\subsubsection{Non-Actionable Attack Culling}
\label{sec:culling}
To improve query efficiency, we first discard non-actionable instances whose labels are unlikely to flip within the perturbation bounds. For a sample $X = [w_1, \dots, w_L]$, we generate $N$ masked variants by replacing each token $w_i$ with "[UNK]" at a fixed probability and compute the retention score $p^0$, estimating the likelihood that the model’s prediction $f(X)$ remains unchanged:
\begin{equation}
p^0 = \frac{1}{N} \sum_{i=1}^{N} \mathds{1}(f(X) = f(X_i))
\end{equation}
Subsequently, we compute the lower bound of $p^0$ using the subsequent equation:
\begin{equation}
p^{0}_{lb}=\min\left \lbrace q\in[0,p^0]: d(p^0,q)<\frac{\beta_{0}}{N}\right \rbrace
\end{equation}
where $\beta_0 = -\log{\delta}$ is an exploration parameter, $d(p,q)$ represents the Kullback-Leibler (KL) divergence between two Bernoulli distributions, with its definition provided by~\citet{pmlr-v30-Kaufmann13}:
\begin{equation}
d(p,q)=p \log\frac{p}{q}+(1-p)\log\frac{1-p}{1-q}
\label{eq:KLD}
\end{equation}

If $p^{0}_{lb}$ exceeds a predefined threshold, modifying the sample’s label is unlikely to alter the output, rendering it non-actionable for adversarial attacks. Such instances are pruned, returning an empty set.

\subsubsection{Construction of Pivot Set}
Our objective is to identify a Pivot Set $S$ that satisfies Eq. \ref{eq:obj}. To keep $|S|$ minimal, we employ an incremental construction strategy. Starting from an empty set ($S^{0} = \emptyset$), each iteration generates candidate sets by adding one word to the current set: $\mathcal{S}^{t} = \{ S^{t-1} \cup \{w_i\} \}_{i=1}^L$. The candidate with the highest estimate retention precision is selected as the updated Pivot Set $S^{t}$. The process  terminates once either $S^{t}$ satisfies Eq. \ref{eq:obj} or a predefined budget limit is reached, returning $S^{t}$ as the final Pivot Set.

To select the optimal Pivot Set at each iteration, we estimate the retention precision of each candidate set. For a candidate $S$, this is computed using samples from $\mathcal{D}(\,\cdot \mid S)$—the distribution of perturbed texts that preserve all words in $S$. To conserve query budget, we aim to minimize the number of samples required. Inspired by~\citet{DBLP:conf/aaai/Ribeiro0G18}, we formulate this as a pure-exploration multi-armed bandit problem: each candidate Pivot Set $S$ is an arm, its true retention precision under $\mathcal{D}(\,\cdot \mid S)$ is the latent reward, and pulling an arm corresponds to querying the model on a sampled instance $z \sim \mathcal{D}(\,z \mid S)$ and checking if the label changes (i.e., evaluating $\mathds{1}\left(f(x) = f(z)\right)$). Under this formulation, we can employ the KL-LUCB algorithm~\citep{pmlr-v30-Kaufmann13} to identify the best Pivot Set $S$. 

According to KL-LUCB, at iteration $t$, the upper and lower confidence bounds of the estimated retention precision $p_{S}$ for Pivot Set $S$ are defined as:
\begin{equation}
\label{eq:p_ub}
p_{S,ub}^t=max\left \lbrace q\in[p_{S},1]: d(p_{S}^t,q)<\frac{\beta(K,t)}{N_{S}^t}\right \rbrace
\end{equation}
\begin{equation}
\label{eq:p_lb}
p_{S,lb}^t=min\left \lbrace q\in[0,p_{S}]: d(p_{S}^t,q)<\frac{\beta(K,t)}{N_{S}^t}\right \rbrace
\end{equation}
where $d(p, q)$ represents the KL divergence between two Bernoulli distributions as defined in Eq. \ref{eq:KLD}, $K$ is the number of arms; in the Pivot Set Selection stage, $K = L - |S|$, where $L$ is the text length. $N_{S}^{t}$ represents the number of times arm $S$ has been pulled before iteration $t$. The exploration parameter $\beta(K,t)$ controls the confidence radius and increases logarithmically with $t$:
\begin{equation}
\beta(K,t) =
\log\left(\frac{\lambda K t^{\alpha}}{\delta}\right)
+ \log\left(\log\left(\frac{\lambda K t^{\alpha}}{\delta}\right)\right)
  \end{equation}
where $\lambda > 0$ is a scaling constant, $\alpha$ determines the growth rate, and $\delta$ is the confidence parameter.

We initialize each rule's estimated retention precision and confidence interval using the sampling procedure in Section \ref{sec:culling}. Pivot Sets are ranked by their estimated retention precision and partitioned into ${\mathcal{S}}^{+}$ and ${\mathcal{S}}^{-}$ based on whether their estimated retention precision exceeds the target threshold $\tau$. To improve estimation accuracy, we iteratively tighten each Pivot Set’s confidence interval. At each iteration, we select the Pivot Set $S \in {\mathcal{S}}^{+}$ with the smallest  estimated retention precision lower bound and $S' \in {\mathcal{S}}^{-}$ with the largest estimated retention precision upper bound, and update both by pulling their corresponding arms.

The sampling process stops when $p_{S,lb}$ exceeds $p_{S', ub}$ within tolerance $\epsilon \in [0,1]$. If $S^{*}$ is the Pivot Set with the highest true retention precision, the following guarantee holds~\citep{pmlr-v30-Kaufmann13}:
\begin{equation}
P(p_{S} \geq p_{S^{*}} - \epsilon) \geq 1 - \delta
\end{equation}

For the Pivot Set $S$ with the highest estimated retention precision in ${\mathcal{S}}^{+}$, we further verify whether it meets the retention precision criterion. If $p_{S,ub} \geq \tau$ but $p_{S,lb} < \tau$, its arm continues to be pulled until we can confidently determine $p_{S,lb} \geq \tau$ (valid Pivot Set) or $p_{S,ub} < \tau$ (invalid Pivot Set).  The process of Pivot Set selection is outlined in Algorithm~\ref{alg:pivot-selection}.

\begin{algorithm}[ht]
\caption{Workflow of Pivot Set Identification}
\begin{algorithmic}[1]
\Function{FindPivot}{$X, \mathcal{D}, \tau$}
    \State \textbf{hyperparameters:} $\epsilon, \delta$
    \State $S^0 \gets \emptyset$
    \Loop
        \State $\mathcal{S}_c^t \gets \textsc{GenerateCands}(\mathcal{S}^{t-1}, X)$
        \State $S^t \gets \textsc{BestCand}(\mathcal{S}_c^t, \mathcal{D}, \epsilon, \delta)$
        \If{$S^t = \emptyset$}
            \State \textbf{break}
        \EndIf
        \If{$p_{S^t} \ge \tau$}
            \State \Return $S^t$
        \EndIf
    \EndLoop
\EndFunction

\vspace{0.5em}

\Function{GenerateCands}{$\mathcal{S}, X$}
    \ForAll{$S \in \mathcal{S}, w_i \in X \setminus S$}
            \State $\mathcal{S} \gets \mathcal{S} \cup \{S \cup \{w_i\}\}$
    \EndFor
    \State \Return $\mathcal{S}$
\EndFunction

\vspace{0.5em}

\Function{BestCand}{$\mathcal{S}, \mathcal{D}, \epsilon, \delta$}
    \State Initialize estimates $p^0$ for all $S \in \mathcal{S}$
    \State Partition $\mathcal{S}$ into $\mathcal{S}^{+}$ and $\mathcal{S}^{-}$ based on $p^0$ 
    \State $S \gets \arg\max_{S \in \mathcal{S}^{+}} p_S$
    \State $S' \gets \arg\max_{S' \in \mathcal{S}^{-}} u_{S'}^t$
    
    \While{$p_{S,ub}^t - p_{S,lb}^t > \epsilon$}
        \State Sample $z \sim \mathcal{D}(z \mid S)$ and $z' \sim \mathcal{D}(z' \mid S')$
        \State Update $p, p_{ub}^t, p_{lb}^t$ for $S$ and $S'$ according to Eq. \ref{eq:p_s}, Eq. \ref{eq:p_ub}, and Eq. \ref{eq:p_lb}
        \State $S \gets \arg\max_{S \in \mathcal{S}^{+}} p_S$
        \State $S' \gets \arg\max_{S' \in \mathcal{S}^{-}} p_{S,ub}^t$
    \EndWhile
    
    \State \Return $S$
\EndFunction
\end{algorithmic}
\label{alg:pivot-selection}
\end{algorithm}

\begin{figure}[!tbp]
  \centering
    \includegraphics[width=\linewidth]
    {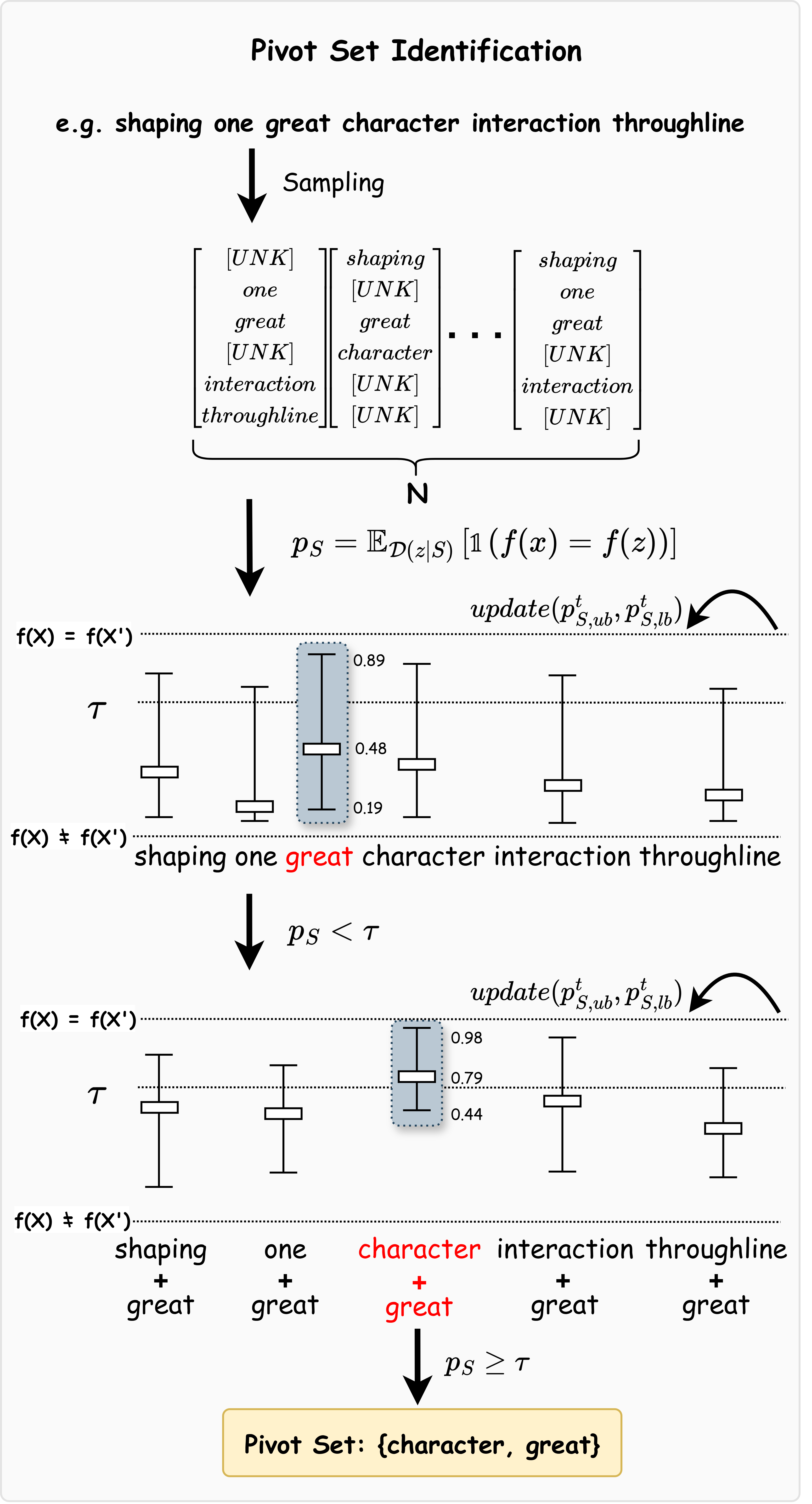}
  \caption{Pivot Set Identification on MR}
  \label{fig:case-study}
\end{figure}

 In addition, we use a real example on the MR dataset to illustrate the Pivot Set selection process (Figure~\ref{fig:case-study}). As shown in the figure, we demonstrate the Pivot Set identification process using the MR example “shaping one great character interaction throughline.” Firstly, the sentence is randomly masked with "[UNK]" tokens to estimate the retention precision $p_{S}$. Using the KL–LUCB procedure, candidate arms are iteratively pulled to tighten their confidence bounds ($u_{i}^t, l_{i}^t$), ensuring that promising tokens are not overlooked. The token “great” initially achieves the highest retention precision, but since $p_{S} < \tau$, an additional token is incorporated. When the pair “great + character” satisfies $p_{S} > \tau$, it is finalized as the Pivot Set.

In {\OURS}, the KL-LUCB component is relatively query-intensive. To prevent excessive budget consumption, {\OURS} allocates it a quota of $\gamma B$, where $\gamma \in [0,1]$ is a user-defined parameter and $B$ is the total query budget. Once this quota is reached, KL-LUCB stops arm-pulling and returns the set $S$ with the highest current estimate retention precision $p$ as the Pivot Set.
If all words in the Pivot Set have been attacked but budget remains, {\OURS} ranks non-pivot words by the retention precision estimates of their corresponding Pivot Set candidates and continues attacking the highest-ranked ones.

It is worth noting that during the construction of the Pivot Set, {\OURS} expands only the candidate set with the highest estimated retention precision, effectively following a greedy strategy. To obtain higher-quality Pivot Sets, one could instead employ beam search, which retains the top-$k$ candidate sets at each step for further expansion—albeit at the cost of a higher query budget.


\subsubsection{Perturbation Execution}
Given an input $X = [w_1, \dots, w_i, \dots, w_L]$ and its Pivot Set $S$, the perturbation stage of {\OURS} consists of three steps: generating substitution candidate sets, selecting adversarial samples, and skipping conspicuous samples.

\noindent\textbf{Generating Substitution Candidate Sets.} 
For each pivot token $w_i \in S$, we construct a substitution candidate set $C(w_i)$ by retrieving the $M$ nearest vocabulary words to $w_i$ in a pretrained embedding space. Concretely, $C(w_i)=\{c_{i,1},\dots,c_{i,M}\}$, where each $c_{i,j}$ is among the $M$ closest words to $w_i$ under cosine similarity. We use counter-fitted word vectors~\citep{mrksic2016counterfitting} to build this space, as they better preserve synonymy and antonymy for lexical substitution.

\noindent\textbf{Selecting Adversarial Samples.}
Replacing $w_i$ with $c_{j}\in C(w_i)$ yields a candidate adversarial sample $X_{j}^{'}=[w_1,\dots,c_{j},\dots,w_L]$
For each pivot token we obtain $M$ candidates. We select the candidate $X_{j}'$ that maximizes cosine similarity to the original input $X$, thereby minimizing semantic drift:
\begin{equation}
X' = \arg\max_{X_j' \in {X_j'}}  \cos\big(\text{embed}(X),\ \text{embed}(X_j')\big)
\end{equation}
where $\text{embed}(\cdot)$ denotes a sentence-level embedding used to measure semantic closeness.

\noindent\textbf{Skipping Conspicuous Samples.}
To preserve stealth, we enforce a perturbation constraint based on the Perturbation Rate:
\begin{equation}
    \text{Pert}(X,X') = \frac{1}{L}\sum_{i=1}^{L}\mathds{1}(w_i \neq w_i')
\end{equation}
where $L$ is the sentence length. Candidates exceeding a threshold $h$ are skipped. We use a dynamic threshold that adapts to the remaining query budget:
\begin{equation}
h = \min\!\left(h_{\max},\, h_{\text{base}} + \frac{B_{\text{rm}}}{L}\right)
\end{equation}
with user-defined $h_{\text{base}}<h_{\max}$ and $B_{\text{rm}}$ denoting remaining queries. This adaptive rule balances stealth and flexibility as budget availability changes.

\section{Experiment}
We aim to investigate the following research questions through our experimental evaluation:
\begin{itemize} [leftmargin=*, noitemsep, topsep=0pt]
    \item  RQ1: How does {\OURS} compare with existing black-box hard-label methods under a limited query budget?
   \item RQ2: How does the performance of {\OURS} vary across different query budgets?
   \item RQ3: How do different NLP tasks influence the effectiveness of {\OURS}?
   \item RQ4: What is the contribution of each component of {\OURS} to its overall performance?
   \item RQ5: How interpretable is {\OURS} relative to other attack models?
\end{itemize}

\subsection{Experimental Setup}
\noindent\textbf{Datasets.} We conducted our experiments on five publicly available text classification datasets: Yelp~\citep{zhang2015character}, Yahoo~\citep{zhang2015character}, MR~\citep{pang2005seeing}, Amazon~\citep{zhang2015character}, and SST-2~\citep{socher2013sst}. To address RQ3, which examines performance across different NLP tasks, we additionally evaluated the attack methods on textual entailment using two datasets: SNLI~\citep{DBLP:conf/emnlp/ConneauKSBB17} and MultiNLI~\citep{williams2018multinli}, the latter of which contains both the matched (MNLI-m) and mismatched (MNLI-mm) splits. The dataset details are provided in Appendix~\ref{app:data}.


\medskip
\noindent\textbf{Victim Models.} Our evaluation covers a broad range of model architectures. Following \citet{DBLP:conf/aaai/ZhuZ0WL24}, we employ WordCNN \citep{Kim_2014}, WordLSTM \citep{Hochreiter_1997}, and BERT \citep{Devlin_2019} as representative classification models. We further include efficient encoder variants, namely ALBERT \citep{DBLP:conf/iclr/LanCGGSS20} and DistilBERT \citep{DBLP:journals/corr/abs-1910-01108}, as well as recent large language models (LLMs): Qwen2.5-1.5B, evaluated in both zero-shot and fine-tuned settings \citep{qwen2_5_techreport_2025}, and Gemma~3~\citep{DBLP:journals/corr/abs-2503-19786}, evaluated in the zero-shot setting . For textual entailment, we adopt BERT. Additional model details are provided in Appendix~\ref{app:models}.


\medskip
\noindent\textbf{Baselines.} We have chosen the following existing hard-label attack algorithms as our baselines: HyGloadAttack~\citep{DBLP:journals/nn/LiuXLYLZX24}, VIWHard~\citep{DBLP:journals/ijon/ZhangWGZWL25}, HLBB~\citep{maheshwary2021hlbb}, TextHoaxer~\citep{ye2022texthoaxer}, LeapAttack~ \citep{DBLP:conf/kdd/YeCMWM22}, TextHacker~\citep{DBLP:conf/emnlp/YuWC022}, and LimeAttack~\citep{DBLP:conf/aaai/ZhuZ0WL24}. More details of the baselines are listed in the Appendix~\ref{app:baselines}.

Details of the evaluation metrics and implementation are provided in Appendix~\ref{appendix:evaluation-metric} and Appendix~\ref{appendix:implementation-detail}, respectively.

\subsection{Overall Result (RQ1)}
\noindent\textbf{Performance Comparison.} Table~\ref{tab:overall} compares {\OURS} against seven baselines under a strict 100-query budget. {\OURS} consistently strikes a superior balance between high ASR and low perturbation across both traditional architectures and LLMs. For instance, on WordLSTM (Yelp), {\OURS} attains 16.8\% ASR (1.4\% Pert) compared to TextHacker's 14.5\% (5.9\% Pert). Against BERT, it achieves 9.7\% ASR with only 1.0\% perturbation, while baselines either lag below 8.2\% ASR or incur higher costs. This dominance extends to LLMs: on Qwen2.5 (Zero-shot/Yahoo), {\OURS} reaches 93.5\% ASR with a mere 1.1\% perturbation, significantly outperforming TextHacker (4.0\% Pert). Even against the robust fine-tuned Qwen2.5, {\OURS} remains the top performer on 4 out of 5 datasets. Additional results on WordCNN and ALBERT are reported in Appendix~\ref{app:more_result}, exhibiting trends consistent with those observed here.

{\OURS}'s superiority stems from two key factors: \textbf{First}, unlike methods like HyGloadAttack or TextHacker that typically start outside from the decision boundary or perform random modifications—often consuming excessive queries to find a valid direction—{\OURS} identifies \textit{pivot words} that dictate the model's prediction and optimizes within the label-invariant region. This inherently reduces query consumption, enabling stronger performance under budget constraints. \textbf{Second}, unlike approaches such as LimeAttack that assess token importance independently (e.g., selecting top-$K$ words in isolation), {\OURS} explicitly models inter-word interactions, thereby capturing the combinatorial effects of multiple edits. This is crucial for both long-text datasets and robust LLMs, where single-word modifications are often insufficient to flip the prediction. 

\begin{table*}[htbp!]
\centering
\scriptsize  
\setlength{\tabcolsep}{2.5pt} 
\renewcommand{\arraystretch}{0.95} 
\resizebox{\textwidth}{!}{
\begin{tabular}{ll cccccccccc}
\toprule
\multirow{2}{*}{\textbf{Model}} & \multirow{2}{*}{\textbf{Attack}} & \multicolumn{2}{c}{\textbf{Yelp}} & \multicolumn{2}{c}{\textbf{Yahoo}} & \multicolumn{2}{c}{\textbf{MR}} & \multicolumn{2}{c}{\textbf{Amazon}} & \multicolumn{2}{c}{\textbf{SST-2}} \\
\cmidrule(lr){3-4} \cmidrule(lr){5-6} \cmidrule(lr){7-8} \cmidrule(lr){9-10} \cmidrule(lr){11-12}
& & \textbf{ASR}$\uparrow$ & \textbf{Pert}$\downarrow$ & \textbf{ASR}$\uparrow$ & \textbf{Pert}$\downarrow$ & \textbf{ASR}$\uparrow$ & \textbf{Pert}$\downarrow$ & \textbf{ASR}$\uparrow$ & \textbf{Pert}$\downarrow$ & \textbf{ASR}$\uparrow$ & \textbf{Pert}$\downarrow$ \\
\midrule
\multirow{8}{*}{\textbf{WordLSTM}} 
& \textbf{PivotAttack} & \textbf{16.8} & \textbf{1.4} & \textbf{42.3} & \textbf{1.5} & \textbf{50.6} & 5.1 & \textbf{18.6} & \textbf{1.8} & \textbf{37.8} & 6.1 \\
& LimeAttack & 11.9 & 2.5 & 39.3 & 2.9 & 48.3 & 5.0 & 18.2 & 2.4 & 36.5 & 5.8 \\
& TextHacker & 14.5 & 5.9 & 38.8 & 4.3 & 44.9 & 5.9 & 18.5 & 3.5 & 33.6 & 6.6 \\
& LeapAttack & 10.9 & 2.6 & 36.4 & 3.0 & 45.2 & 4.9 & 15.0 & 2.2 & 33.4 & 5.6 \\
& TextHoaxer & 9.5 & 2.3 & 34.4 & 3.8 & 42.8 & \textbf{4.8} & 14.4 & 2.5 & 29.2 & \textbf{5.5} \\
& HLBB & 11.1 & 2.7 & 37.9 & 3.0 & 41.7 & 5.4 & 17.6 & 3.1 & 27.0 & 5.9 \\
& VIWHard & 13.3 & 1.5 & 41.9 & 1.6 & 47.3 & 5.2 & 14.7 & 1.9 & 35.3 & 6.1 \\
& HyGloadAttack & 11.7 & 2.3 & 37.9 & 3.4 & 46.2 & 5.3 & 17.8 & 2.4 & 33.5 & 6.4 \\
\midrule

\multirow{8}{*}{\textbf{BERT}} 
& \textbf{PivotAttack} & \textbf{9.7} & \textbf{1.0} & \textbf{39.7} & \textbf{1.9} & \textbf{47.5} & 5.9 & 15.5 & 2.2 & \textbf{30.6} & 6.4 \\
& LimeAttack & 7.2 & 2.6 & 36.6 & 3.0 & 40.1 & 5.8 & 13.4 & 2.5 & 27.2 & 6.1 \\
& TextHacker & 7.4 & 2.5 & 34.6 & 3.9 & 38.5 & 6.2 & 14.8 & 3.6 & 19.2 & 6.9 \\
& LeapAttack & 6.1 & 2.6 & 32.2 & 2.9 & 37.6 & 5.0 & 12.0 & 2.6 & 22.5 & 6.2 \\
& TextHoaxer & 5.8 & 3.4 & 29.6 & 2.9 & 37.0 & \textbf{4.9} & 11.1 & 2.7 & 18.7 & \textbf{5.7} \\
& HLBB & 6.4 & 2.0 & 35.0 & 2.7 & 36.6 & 5.3 & 13.7 & 3.1 & 17.8 & 5.9 \\
& VIWHard & 8.2 & 1.2 & 39.0 & 2.0 & 36.3 & 5.5 & 11.4 & \textbf{1.9} & 26.1 & 6.8 \\
& HyGloadAttack & \textbf{9.7} & 2.1 & 36.0 & 3.4 & 41.3 & 5.7 & \textbf{16.3} & 3.3 & 25.0 & 6.7 \\
\midrule

\multirow{8}{*}{\textbf{DistilBERT}} 
& \textbf{PivotAttack} & \textbf{8.6} & \textbf{1.2} & \textbf{37.1} & 1.8 & \textbf{33.5} & 6.2 & \textbf{11.7} & \textbf{1.5} & \textbf{32.4} & 6.5 \\
& LimeAttack & 8.0 & 2.4 & 36.1 & 2.3 & 28.7 & 5.4 & 10.8 & 1.6 & 27.8 & \textbf{6.0} \\
& TextHacker & 8.0 & 2.3 & 35.4 & 2.6 & 27.1 & 6.0 & 10.5 & 2.0 & 27.1 & 6.6 \\
& LeapAttack & 7.6 & 2.6 & 33.4 & 3.7 & 24.3 & 5.9 & 10.5 & 2.4 & 25.9 & 6.8 \\
& TextHoaxer & 7.0 & 2.1 & 32.2 & 3.5 & 21.1 & \textbf{5.5} & 9.5 & 2.5 & 20.0 & 6.6 \\
& HLBB & 6.8 & 2.1 & 31.2 & 3.7 & 17.0 & 6.0 & 10.9 & 3.2 & 16.2 & 6.8 \\
& VIWHard & 6.7 & 1.3 & 36.8 & \textbf{1.7} & 21.9 & 5.7 & 8.3 & 1.6 & 27.6 & 7.0 \\
& HyGloadAttack & 8.4 & 2.0 & 33.5 & 3.6 & 26.2 & 5.7 & 11.5 & 2.4 & 24.4 & 6.8 \\
\midrule

\multirow{8}{*}{\shortstack{\textbf{Gemma 3}\\\textbf{(Zero-shot)}}} 
& \textbf{PivotAttack} & \textbf{39.8} & \textbf{0.6} & \textbf{88.7} & 1.0 & \textbf{62.2} & 4.8 & \textbf{34.9} & \textbf{1.2} & \textbf{54.6} & 5.9 \\
& LimeAttack & 38.6 & 0.8 & 84.1 & \textbf{0.8} & 51.2 & 4.9 & 34 & 1.3 & 53.4 & \textbf{5.5} \\
& TextHacker & 34.1 & 2.1 & 85.5 & 1.6  & 59.6 & 5.0 & 33.6 & 3.5 & 51.2 & 6.1  \\
& LeapAttack & 29.8 & 2.4 & 83.2 & 2.1 & 60.3 & 4.6 & 33.2 & 2.1 & 50.8 & 6.0 \\
& TextHoaxer & 25.5 & 2.3 & 82.8 & 1.8 & 53.5 & 4.3 & 32.1 & 2.4 & 50.3 & 6.2 \\
& HLBB & 23.3 & 8.2 & 83.1 & 2.5 & 52.6 & 5.0 & 32.6 & 4.5 & 49.2 & 6.4 \\
& VIWHard & 33.1 & 1.0 & 86.1 & 1.3 & 56.9 & \textbf{4.2} & 32.7 & 3.2 & 53.1 & 6.3 \\
& HyGloadAttack & 35.8 & 2.1 & 87.5 & 2.5 & 60.1 & 4.7 & 33.4 & 3.9 & 53.8 & 6.7 \\
\midrule

\multirow{8}{*}{\shortstack{\textbf{Qwen2.5}\\\textbf{(Zero-shot)}}} 
& \textbf{PivotAttack} & \textbf{16.1} & \textbf{1.0} & \textbf{93.5} & \textbf{1.1} & \textbf{44.1} & 6.4 & 21.8 & \textbf{1.7} & \textbf{52.6} & 6.1 \\
& LimeAttack & 13.5 & 2.2 & 92.7 & 2.5 & 43.1 & 5.3 & \textbf{23.7} & 3.0 & 49.6 & \textbf{5.5} \\
& TextHacker & 14.1 & 3.7 & 93.1 & 4.0 & 42.8 & 6.9 & 22.8 & 4.3 & 45.2 & 6.2 \\
& LeapAttack & 13.9 & 2.9 & 92.8 & 3.0 & 42.0 & \textbf{5.1} & 20.8 & 3.9 & 41.4 & 5.7 \\
& TextHoaxer & 12.5 & 3.1 & 91.3 & 3.1 & 41.1 & 6.4 & 21.2 & 3.7 & 42.5 & 6.0 \\
& HLBB & 13.9 & 2.7 & 82.7 & 2.8 & 35.9 & 5.3 & 18.6 & 3.2 & 37.0 & 5.9 \\
& VIWHard & 14.1 & 1.3 & 89.2 & 1.3 & 36.9 & 6.0 & 19.2 & 2.1 & 48.2 & 6.6 \\
& HyGloadAttack & 13.6 & 2.0 & 83.7 & 1.9 & 35.1 & 5.4 & 17.6 & 2.5 & 45.9 & 6.7 \\
\midrule

\multirow{8}{*}{\shortstack{\textbf{Qwen2.5}\\\textbf{(Fine-tuned)}}} 
& \textbf{PivotAttack} & \textbf{3.9} & \textbf{1.6} & \textbf{46.2} & \textbf{1.2} & \textbf{32.1} & 6.4 & \textbf{6.5} & \textbf{2.0} & \textbf{29.6} & 6.4 \\
& LimeAttack & 3.7 & 2 & 41.3 & 1.6 & 30.0 & 5.5 & 6.0 & 2.5 & 25.6 & \textbf{6.1} \\
& TextHacker & 3.7 & 2.3 & 45.5 & 1.8 & 29.6 & 6.0 & 6.0 & 2.2 & 26.4 & 6.2 \\
& LeapAttack & 3.4 & 2.2 & 42.7 & 2.1 & 28.3 & 5.7 & 6.3 & 2.5 & 25.1 & 6.3 \\
& TextHoaxer & 3.2 & 2.5 & 40.8 & 1.8 & 28.1 & \textbf{5.4} & 5.9 & 2.8 & 24.5 & 6.7 \\
& HLBB & 3.6 & 3.1 & 41.1 & 2.4 & 24.5 & 5.6 & 5.7 & 2.4 & 26.8 & 6.3 \\
& VIWHard & 2.9 & 2.3 & 44.3 & 1.4 & 26.4 & 6.0 & 5.8& 2.2 & 24.8 & 6.4 \\
& HyGloadAttack & 3.6 & 2.4 & 45.8 & 2.9 & 29.1 & 5.8 & 6.1 & 2.9 & 25.3 & 6.9 \\

\bottomrule
\end{tabular}

}
\caption{Comprehensive performance comparison (ASR \% and Pert \%) across all victim models and datasets under a 100-Query Budget. }
\label{tab:overall}
\end{table*}

\medskip
\noindent\textbf{Adversary Quality.} We further evaluate the quality of adversarial examples on the Yelp dataset against BERT, specifically measuring semantic similarity and grammatical error rate. These results are detailed in Appendix~\ref{appendix:adversary-quality}.

\subsection{Query Budget (RQ2)}
\begin{figure}[htbp]
  \centering
  \includegraphics[width=0.8\linewidth]{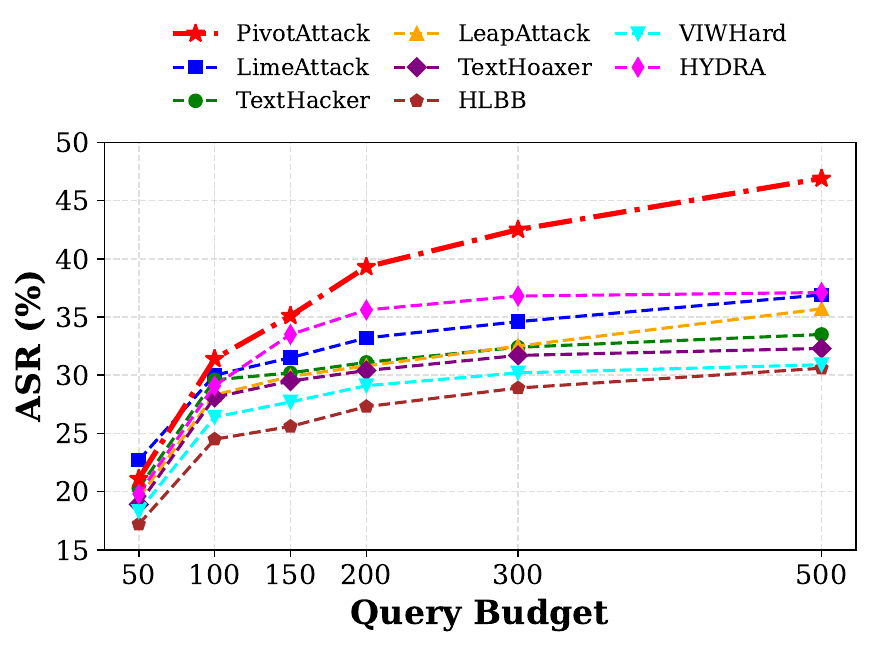}
  \caption{ASR vs. Query Budget: MR (Qwen2.5-FT)}
\label{pic:query-budget}
\end{figure}

We evaluate different attack methods on the MR and SST-2 datasets under varying query budgets (B = 50, 100, 150, 200, 300, 500). Due to space limitations, only MR (Qwen2.5-FT) results are shown in Figure~\ref{pic:query-budget}, with full results in Appendix~\ref{appendix:query-budget}. As shown in the figure, {\OURS} consistently outperforms other methods in ASR as the query budget increases, with its advantage even becoming more pronounced. This is because a larger query budget allows the KL-LUCB component to perform more arm pulls, yielding more accurate retention precision estimates and thus identifying a better Pivot Set.


\subsection{Transferability (RQ3)}
\begin{table}[t]
\centering
\small
\setlength{\tabcolsep}{3.5pt}
\begin{adjustbox}{max width=\linewidth}
\begin{tabular}{l cc cc cc}
\toprule
\multirow{2}{*}{\textbf{Attack}}
 & \multicolumn{2}{c}{\textbf{SNLI}}
 & \multicolumn{2}{c}{\textbf{MNLI-m}}
 & \multicolumn{2}{c}{\textbf{MNLI-mm}}\\
\cmidrule(lr){2-3}\cmidrule(lr){4-5}\cmidrule(lr){6-7}
 & {\textbf{ASR.}$\uparrow$} & {\textbf{Pert.}$\downarrow$}
 & {\textbf{ASR.}$\uparrow$} & {\textbf{Pert.}$\downarrow$}
 & {\textbf{ASR.}$\uparrow$} & {\textbf{Pert.}$\downarrow$} \\
\midrule
\textbf{{\OURS}} & \textbf{25.8} & 7.9 & \textbf{47.7} & 7.7 & \textbf{54.8} & 8.0 \\
LimeAttack            & 25.3 & 8.0 & 46.4 & 7.2 & 52.3 & 7.2 \\
TextHacker            & 22.1 & 8.3 & 38.2 & 7.7 & 44.1 & 7.5 \\
LeapAttack            & 24.4 & \textbf{7.6} & 43.1 & 7.6 & 45.9 & 7.6 \\
TextHoaxer            & 23.6 & 7.7 & 42.2 & 7.2 & 46.8 & \textbf{7.0} \\
HLBB                  & 23.6 & 7.9 & 43.6 & \textbf{7.1} & 46.3 & \textbf{7.0} \\
VIWHard               & 24.1 & 7.8 & 45.1 & 7.6 & 53.3 & 7.3 \\
HyGloadAttack         & 25.5 & 8.8 & 46.5 & 7.7 & 54.0 & 7.5 \\
\bottomrule
\end{tabular}
\end{adjustbox}
\caption{Textual Entailment Attack Performance on BERT (100-Query Budget)}
\label{tab:entail}
\end{table}

To evaluate the performance of different text adversarial attack algorithms across NLP tasks, we conducted experiments on the textual entailment datasets SNLI and MNLI. The results, shown in Table \ref{tab:entail}, indicate that {\OURS} achieves the highest ASR while maintaining perturbation levels comparable to other methods.


\subsection{Ablation Study (RQ4)}
\noindent\textbf{Ablation Study of {\OURS}.} We conduct an ablation study on Yelp and MR (LSTM, 100 queries) with three variants: (1) randomizing the Pivot Set (\textit{$-$Pivot Set}); (2) fixing the perturbation threshold $h$ at 0.1 (\textit{$-$Dynamic constraints}); and (3) randomly ranking non-pivot words instead of reusing the intermediate KL-LUCB output (\textit{$-$Reuse}). Table~\ref{tab:ablation-attack} shows that while perturbation rates remain stable, \textit{$-$Pivot Set} causes the largest ASR drop (e.g., 16.8\%\(\to\)13.7\% on Yelp), confirming the efficacy of pivot targeting. \textit{$-$Dynamic constraints} also reduces performance, whereas \textit{$-$Reuse} has minimal impact, suggesting that attacking pivot words alone is typically sufficient.



\begin{table}[ht]\centering\small
\begin{adjustbox}{max width=\linewidth}
\begin{tabular}{llcc}
\toprule
\textbf{Dataset} & \textbf{Method} & \textbf{ASR.}$\uparrow$ & \textbf{Pert.}$\downarrow$ \\
\midrule
\multirow{4}{*}{\textbf{Yelp}}
 & \textbf{\OURS} & \textbf{16.8} & \textbf{1.4} \\
 & \textminus{}Pivot Set & 13.7 & \textbf{1.4} \\
 & \textminus{}Dynamic constraints & 14.4 & \textbf{1.4} \\
 & \textminus{}Reuse & 16.6 & \textbf{1.4} \\
\midrule
\multirow{4}{*}{\textbf{MR}}
 & \textbf{\OURS} & \textbf{50.6} & 5.1 \\
 & \textminus{}Pivot Set & 46 & 5.2 \\
 & \textminus{}Dynamic constraints & 47.2 & \textbf{4.8} \\
 & \textminus{}Reuse & 50.2 & 5.2 \\
\midrule
\end{tabular}
\end{adjustbox}
\caption{Ablation study of {\OURS} components.}
\label{tab:ablation-attack}
\end{table}



\noindent\textbf{Ablation Study of Pviot Set Identification.} We evaluate component contributions under a 300-query budget using three variants: (1) skipping attack culling (\textit{$-$Culling}), (2) omitting multi-armed bandit refinement (\textit{$-$MAB}), and (3) disabling retention precision tightening (\textit{$-$Tighten}). As shown in Table~\ref{tab:ablation-set}, \textit{$-$MAB} causes the largest performance drop, confirming the necessity of retention precision refinement. \textit{$-$Culling} also reduces efficacy by wasting queries, while \textit{$-$Tighten} has negligible impact as it is rarely triggered.



\begin{table}[ht]\centering\small
\label{tab:ablation-set}
\begin{tabular}{llcc}
\toprule
\textbf{Dataset} & \textbf{Method} & \textbf{ASR.} & \textbf{Pert.} \\
\midrule
\multirow{4}{*}{\textbf{Yelp}}
 & \textbf{{\OURS}} & \textbf{15.2} & \textbf{1.2} \\
 & \textminus{}Culling & 13.8 & 1.4 \\
 & \textminus{}MAB  & 13.6 & 1.4 \\
 & \textminus{}Tighten & 14.8 & 1.3 \\
\midrule
\multirow{4}{*}{\textbf{MR}}
 & \textbf{{\OURS}} & \textbf{49.8} & \textbf{5.0} \\
 & \textminus{}Culling & 47.9 & \textbf{5.0} \\
 & \textminus{}MAB  & 46.3 & 5.1 \\
 & \textminus{}Tighten & 49.5 & 5.1 \\
\bottomrule
\end{tabular}
\caption{Ablation study of the Pivot Set Identification module.}
\label{tab:ablation-set}
\end{table}



\subsection{Human Evaluation (RQ5)}
\begin{figure}[!tbp]
  \centering
  \begin{subfigure}{0.49\linewidth}
    \includegraphics[width=\linewidth]{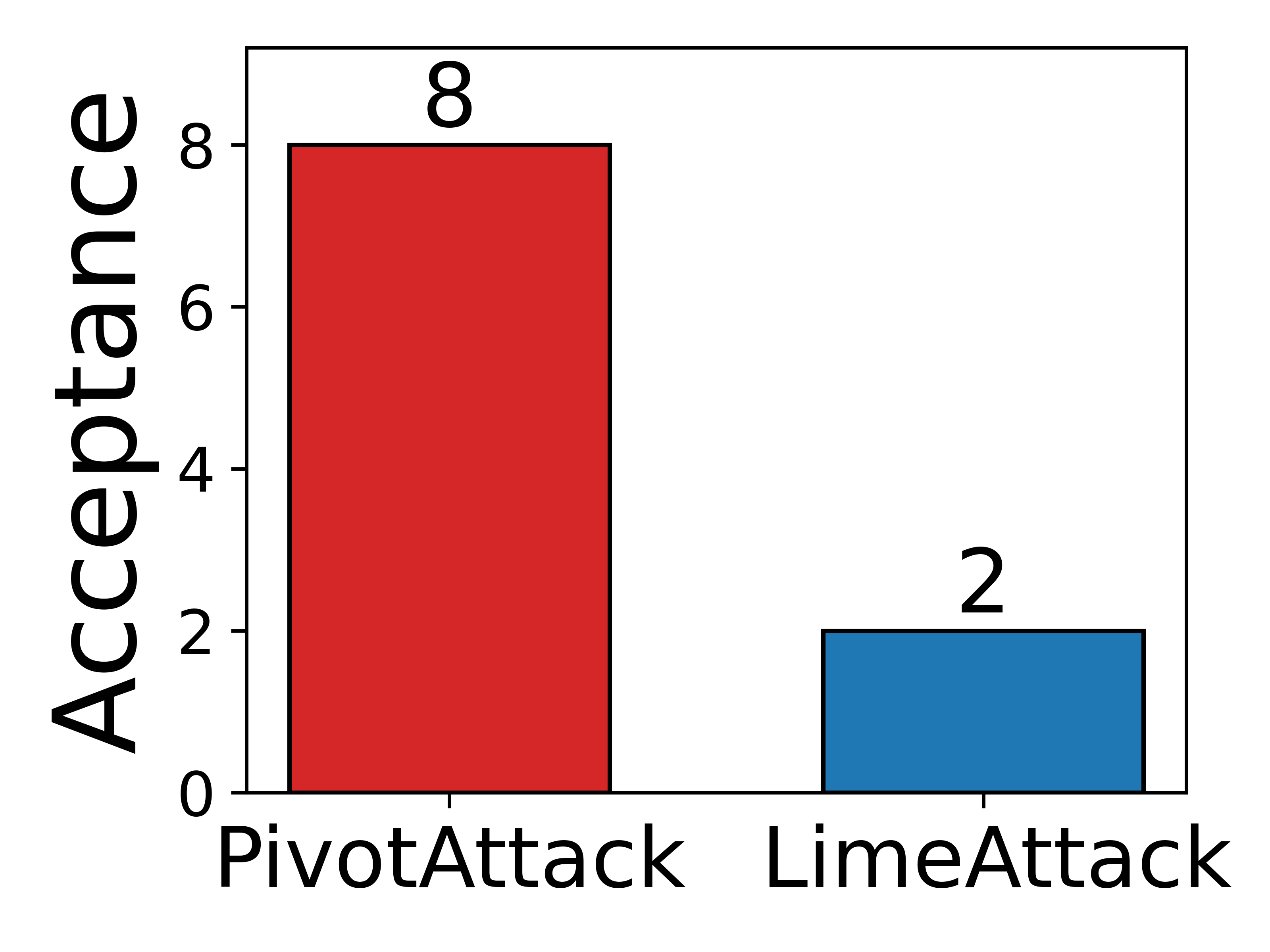}
    \caption{Reasonableness}
  \end{subfigure}\hfill
  \begin{subfigure}{0.49\linewidth}
    \includegraphics[width=\linewidth]{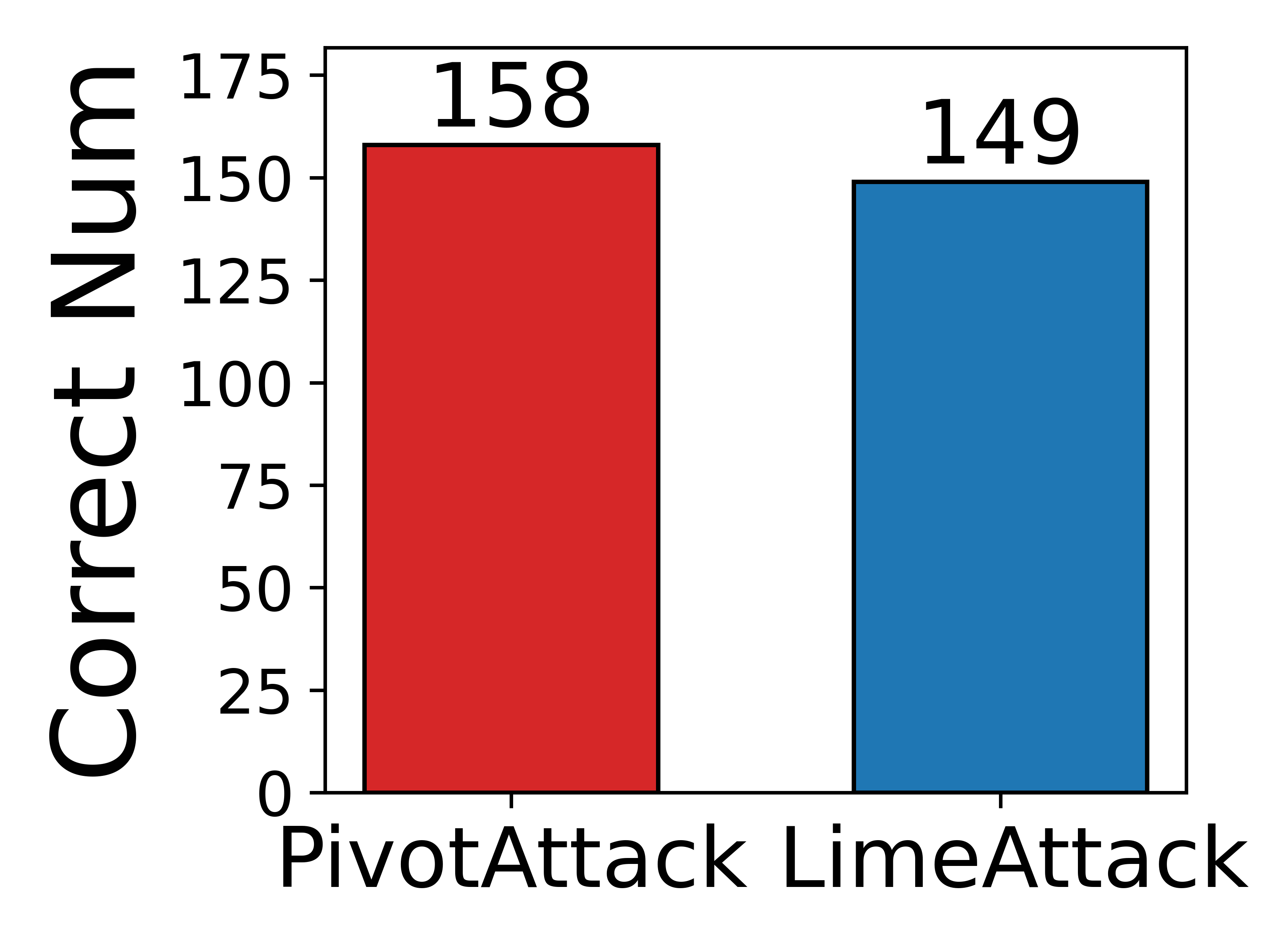}
    \caption{Prediction}
  \end{subfigure}\hfill
  \caption{Human Evaluation}
\label{fig:human-evaluatation}
\end{figure}

To assess the interpretability of {\OURS}, we conducted a human study with 10 participants, comparing it against the interpretable baseline, LimeAttack. We examined which algorithm’s selected perturbation words were more predictive and more reasonable to humans.
For each method, participants first reviewed 10 sentences with the top-2 important words identified by the algorithm, then answered 20 multiple-choice questions based on SST-2 sentences. Each question included the top-2 words from both algorithms among five options, allowing us to evaluate participants’ understanding of each model’s word-importance rationale. Finally, participants selected which method produced more reasonable results. As shown in Figure~\ref{fig:human-evaluatation}, {\OURS} achieved slightly higher predictability and was generally judged more reasonable. 

Qualitatively, participants observed that LimeAttack often prioritized functional words (e.g., "of"), whereas {\OURS} targeted semantically meaningful tokens. For instance, in the example "It's hard to resist his enthusiasm" (Figure~\ref{fig:human-combined}, Appendix~\ref{app:human}), {\OURS} identifies the Pivot Set \{hard, resist\}, correctly locating the semantic anchor where modification destroys the positive sentiment. In contrast, LimeAttack highlights trivial tokens such as \{even, it\}, which may trigger statistical fluctuations in the model but lack genuine semantic significance.

\section{Conclusion}
We presented {\OURS}, a query-efficient hard-label black-box attack that fundamentally shifts the paradigm from approximating decision boundaries to identifying Pivot Sets—combinatorial tokens that act as the "load-bearing walls" of a model's prediction. By perturbing these pivot words, it efficiently drives inputs toward the decision boundary. Experiments show that {\OURS} consistently outperforms baselines across datasets and victim models under limited query budgets, demonstrating the effectiveness of targeting pivot words and modeling inter-word dependencies. Notably, it exposes the vulnerability of Large Language Models (e.g., Qwen2.5, Gemma 3) in both zero-shot and robust fine-tuned settings, achieving high success rates with minimal perturbation. 

\section*{Limitation}
Although {\OURS} outperforms baselines under limited query budgets, the KL-LUCB component of the multi-armed bandit used for Pivot Set identification is relatively query-intensive. As a result, we currently rely on a greedy search to select the best Pivot Set under constrained budgets, which prevents the use of more advanced strategies, such as beam search, for potentially better pivot selection. In future work, we plan to investigate methods to reduce the query cost of the multi-armed bandit component.

\bibliography{custom}

\clearpage
\appendix

\section{Dataset Details}
\label{app:data}
Our experiments utilized five text classification datasets: Yelp~\citep{zhang2015character}, Yahoo~\citep{zhang2015character}, MR~\citep{pang2005seeing}, Amazon~\citep{zhang2015character}, and SST-2~\citep{socher2013sst}. Consistent with prior work on text adversarial attacks~\citep{DBLP:conf/aaai/ZhuZ0WL24,maheshwary2021hlbb}, we constructed our test sets by sampling 1000 examples from each corpus. Specifically, for Yelp, Yahoo, and MR, we utilized the publicly available datasets released by HLBB. For Amazon and SST-2, our samples were drawn randomly from their respective corpora. Notably, certain entries in the SST-2 dataset contained very few tokens, making them unsuitable for attack within our Pert budget; consequently, we restricted our sampling to sentences exceeding 10 tokens in length. Details regarding these datasets are provided below:
\begin{itemize}
  \item \textbf{Yelp}: User reviews annotated with binary sentiment labels (positive/negative).
  \item \textbf{Yahoo}: Community QA dataset for topic classification across 10 categories: Society \& Culture, Science \& Mathematics, Health, Education \& Reference, Computers \& Internet, Sports, Business \& Finance, Entertainment \& Music, Family \& Relationships, Politics \& Government.
  \item \textbf{MR}: Binary sentiment classification task on movie reviews using the standard train/test split.
  \item \textbf{Amazon}: E-commerce review dataset; we use the binary polarity version (positive/negative).
  \item \textbf{SST-2}: Stanford Sentiment Treebank (SST) version adapted for binary classification at the sentence level.
\end{itemize}

Our evaluation on textual entailment utilizes two benchmark datasets: SNLI~\citep{DBLP:conf/emnlp/ConneauKSBB17} and MultiNLI~\citep{williams2018multinli}, with the latter encompassing both matched (MNLI-m) and mismatched (MNLI-mm) partitions. The experimental data for both datasets were sourced from the public repository accompanying the HLBB implementation. Details of each dataset are provided below:
\begin{itemize}
\item \textbf{SNLI:} Consists of human-annotated premise-hypothesis pairs categorized into three classes: entailment, contradiction, and neutrality.
\item \textbf{MultiNLI:} Represents a comprehensive NLI resource offering two distinct evaluation settings—a matched domain (MNLI-m) and a challenging mismatched domain (MNLI-mm).
\end{itemize}

The dataset statistics are summarized in Table~\ref{tab:dataset}.

\begin{table}\centering\small
\begin{adjustbox}{max width=\linewidth}
\begin{tabular}{l|lcccc}
\toprule
\textbf{Task} & \textbf{Dataset} & \textbf{Train} & \textbf{Test} & \textbf{Class} & \textbf{Length} \\
\midrule
\multirow{5}{*}{\textbf{Classification}}
 & Yelp   & 560K  & 38K  & 2  & 133 \\
 & Yahoo  & 1400K & 60K  & 10 & 151 \\
 & MR     & 9K    & 1K   & 2  & 18  \\
 & Amazon & 3600K & 400K & 2  & 79  \\
 & SST-2  & 70K   & 2K   & 2  & 8   \\
\midrule
\multirow{2}{*}{\textbf{Entailment}}
 & SNLI        & 570K & 3K  & 3 & 20 \\
 & MNLI  & 433K & 10K & 3 & 11 \\
\bottomrule
\end{tabular}
\end{adjustbox}
\caption{Dataset Statistics}
\label{tab:dataset}
\end{table}

\section{Victim Model Details}
\label{app:models}
We evaluate our attack against a diverse set of victim models spanning three architectural paradigms: traditional deep learning models, including WordCNN~\citep{Kim_2014} and WordLSTM~\citep{Hochreiter_1997}; pre-trained language models (PLMs) such as BERT~\citep{Devlin_2019}, DistilBERT~\citep{DBLP:journals/corr/abs-1910-01108}, and ALBERT~\citep{DBLP:conf/iclr/LanCGGSS20}; and large language models (LLMs), specifically Gemma~3~\citep{DBLP:journals/corr/abs-2503-19786} and Qwen~2.5~\citep{qwen2_5_techreport_2025}. Their specific architectural details are outlined below:

\begin{itemize}
    \item \textbf{WordCNN}: A convolutional neural network utilizing 200-dimensional GloVe embeddings (trained on 6B tokens). It consists of three convolutional kernels (sizes 3, 4, and 5) with 100 filters each, followed by a dropout layer ($p=0.3$).
    
    \item \textbf{WordLSTM}: A bidirectional LSTM with 150 hidden units, employing the same embedding initialization and regularization settings as WordCNN.
    
    \item \textbf{BERT}: The standard BERT-base architecture (12 layers, 768 hidden units, 12 attention heads). Input sequences are fixed to 256 tokens via padding or truncation.
    
    \item \textbf{DistilBERT}: A distilled version of BERT comprising six layers, each with 768 hidden units and 12 attention heads. We utilize the final-layer \texttt{[CLS]} token representation for classification, applying dropout for regularization.
    
    \item \textbf{ALBERT}: A parameter-efficient variant comprising 12 layers (768 hidden units, 12 attention heads) with factorized embeddings (projecting 128-dim to 768-dim) and cross-layer parameter sharing. We use a fixed sequence length of 256 and feed the final \texttt{[CLS]} representation into a linear classification head.
    
    \item \textbf{Gemma 3 Zero-shot (Gemma 3-ZS)}: We evaluate the Gemma 3-1B model in a zero-shot setting.   The specific prompts used for classification are illustrated in Figure~\ref{fig:prompts}.
    
    \item \textbf{Qwen2.5 Zero-shot (Qwen2.5-ZS)}: The Qwen2.5-1.5B model evaluated in a zero-shot configuration, using the same prompt structure shown in Figure~\ref{fig:prompts}. 
    
    \item \textbf{Qwen2.5 Fine-tuned (Qwen2.5-FT)}: Task-specific variants of Qwen2.5-1.5B obtained via QLoRA fine-tuning. The base model parameters are frozen, and LoRA adapters ($r=16, \alpha=32, \text{dropout}=0.05$) are optimized on the query, key, value, and output projections. Classification is cast as a generation task outputting a single numeric label via instruction-style prompts (see Figure~\ref{fig:prompts}), with cross-entropy loss applied exclusively to the target token. 

\end{itemize}

The original accuracy of the victim models on various classification and entailment datasets is reported in Table~\ref{tab:ori-acc}\footnote{Large language models (e.g., Qwen and Gemma) may refuse to generate responses for certain inputs. Consequently, instances triggering such refusals were excluded from the evaluation for these models.}.

\begin{table}[t]
\centering
\small

\begin{subtable}[t]{\columnwidth}
\centering
\label{tab:ori-acc-a}
\resizebox{\columnwidth}{!}{%
\begin{tabular}{lccccc}
\toprule
Model & Yelp & Yahoo & MR & Amazon & SST-2 \\
\midrule
CNN        & 93.6 & 71.1 & 76.5 & 91.1 & 86.4 \\
LSTM       & 94.6 & 73.7 & 78.0 & 92.2 & 87.8 \\
BERT       & 96.5 & 79.1 & 85.0 & 94.2 & 98.6 \\
DistilBERT & 96.3 & 78.9 & 97.6 & 96.1 & 98.6 \\
ALBERT     & 93.2 & 78.3 & 93.0 & 93.2 & 95.3 \\
Gemma 3-ZS     & 81.7 & 19.6 & 71.5 & 79.7 & 61.0 \\
Qwen2.5-ZS    & 89.0 & 23.7 & 79.7 & 88.5 & 85.3 \\
Qwen2.5-FT & 98.3 & 64.0 & 97.0 & 97.1 & 95.8 \\
\bottomrule
\end{tabular}}
\caption{Text Classification Datasets}
\end{subtable}

\begin{subtable}[t]{\columnwidth}
\centering
\label{tab:ori-acc-b}
\begin{tabular}{lccc}
\toprule
Model & SNLI & MNLI-m & MNLI-mm \\
\midrule
BERT  & 89.1 & 85.1   & 82.2 \\
\bottomrule
\end{tabular}
\caption{Natural Language Inference Datasets}
\end{subtable}


\caption{Original Accuracy of the Victim Models on Classification and NLI Tasks}
\label{tab:ori-acc}
\end{table}

\begin{figure}
\centering
\begin{subfigure}{\linewidth}
\begin{tcolorbox}[title=2 classes prompt]
\footnotesize\ttfamily\raggedright
Help me determine whether the following sentence is negative or positive. \\
Answer with 0 for negative and 1 for positive, and only respond with 0 or 1.
\end{tcolorbox}
\caption{Yelp (2 classes)}
\end{subfigure}\hfill
\begin{subfigure}{\linewidth}
\begin{tcolorbox}[title=10 classes prompt]
\scriptsize\ttfamily\raggedright
You are a strict Yahoo Answers topic classifier. Return ONLY one digit 0-9 with NO other text. \\
Mapping: 0=Society \& Culture, 1=Science \& Mathematics, 2=Health... \\
Examples: \\
Q: "Would my girlfriend break up ..." -> 0 \\
Q: "Whats an easy way to visualize..." -> 1 \\
Q: "when I'm trying too lose..." -> 2 \\
... \\
Answer:
\end{tcolorbox}
\caption{Yahoo (10 classes)}
\end{subfigure}
\caption{Zero-shot LLM prompts used in hard-label evaluation on different classes. It shows the exact prompts we use during evaluation. }
\label{fig:prompts}
\end{figure}

\section{Baseline Details} 
\label{app:baselines}

We compare {\OURS} against five representative hard-label attack algorithms:

\begin{itemize}
\item \textbf{HLBB~\citep{maheshwary2021hlbb}}: A population-based genetic algorithm that optimizes adversarial examples by iteratively selecting candidates with high semantic similarity and low perturbation under a strict query budget.

\item \textbf{TextHoaxer~\citep{ye2022texthoaxer}}: A greedy heuristic that prioritizes token positions, using a few probing queries to determine whether a substitution is worthwhile before committing.

\item \textbf{LeapAttack~\citep{DBLP:conf/kdd/YeCMWM22}}: Employs finite-difference boundary exploration with gradient-like directions to jointly select positions and synonyms, pushing inputs into the misclassification region.

\item \textbf{TextHacker~\citep{DBLP:conf/emnlp/YuWC022}}: Maintains an online-updated word-importance table derived from edit-flip history, which guides a hybrid local search.

\item \textbf{LimeAttack~\citep{DBLP:conf/aaai/ZhuZ0WL24}}: Leverages a local explainable method to approximate word importance ranking,
and then adopts beam search to find the optimal solution.

\item \textbf{VIWHard~\citep{DBLP:journals/ijon/ZhangWGZWL25}}: Introduces an important-word discriminator trained on a local surrogate model to identify vulnerable tokens without querying the target. It generates context-aware substitutions via a Masked Language Model (MLM) and optimizes the attack using a genetic algorithm.


\item \textbf{HyGloadAttack~\citep{DBLP:journals/nn/LiuXLYLZX24}}: Integrates gradient-based search in the embedding space with exchange-based mechanisms via a hybrid optimization strategy. It employs global random initialization to escape local optima and accelerate the search process.


\end{itemize}

\begin{table*}[t]
\centering
\scriptsize  
\setlength{\tabcolsep}{2.5pt} 
\renewcommand{\arraystretch}{0.95} 
\resizebox{\textwidth}{!}{
\begin{tabular}{ll cccccccccc}
\toprule
\multirow{2}{*}{\textbf{Model}} & \multirow{2}{*}{\textbf{Attack}} & \multicolumn{2}{c}{\textbf{Yelp}} & \multicolumn{2}{c}{\textbf{Yahoo}} & \multicolumn{2}{c}{\textbf{MR}} & \multicolumn{2}{c}{\textbf{Amazon}} & \multicolumn{2}{c}{\textbf{SST-2}} \\
\cmidrule(lr){3-4} \cmidrule(lr){5-6} \cmidrule(lr){7-8} \cmidrule(lr){9-10} \cmidrule(lr){11-12}
& & \textbf{ASR}$\uparrow$ & \textbf{Pert}$\downarrow$ & \textbf{ASR}$\uparrow$ & \textbf{Pert}$\downarrow$ & \textbf{ASR}$\uparrow$ & \textbf{Pert}$\downarrow$ & \textbf{ASR}$\uparrow$ & \textbf{Pert}$\downarrow$ & \textbf{ASR}$\uparrow$ & \textbf{Pert}$\downarrow$ \\
\midrule

\multirow{8}{*}{\textbf{WordCNN}} 
& \textbf{PivotAttack} & \textbf{12.9} & \textbf{1.6} & \textbf{44.5} & \textbf{2.4} & \textbf{51.5} & 6.0 & \textbf{19.0} & \textbf{2.1} & \textbf{46.4} & 6.1 \\
& LimeAttack & 11.1 & 2.3 & 43.2 & 2.9 & 48.8 & 5.1 & 17.0 & 2.4 & 43.6 & 5.8 \\
& TextHacker & \textbf{12.9} & 3.0 & 43.0 & 4.6 & 48.6 & 6.1 & \textbf{19.0} & 2.9 & 37.2 & 6.6 \\
& LeapAttack & 10.0 & 2.6 & 38.1 & 2.7 & 46.7 & 4.9 & 16.1 & 2.3 & 41.7 & 5.5 \\
& TextHoaxer & 9.8 & 2.6 & 38.0 & 2.8 & 42.3 & \textbf{4.7} & 16.0 & 2.5 & 37.5 & \textbf{5.3} \\
& HLBB & 11.9 & 2.9 & 39.1 & 3.0 & 41.3 & 5.3 & 16.8 & 2.9 & 37.3 & 5.7 \\
& VIWHard & 10.7 & 1.7 & 42.5 & 2.6 & 47.5 & 5.2 & 14.7 & 2.3 & 39.7 & 6.1 \\
& HyGloadAttack & 11.3 & 2.7 & 40.9 & 3.5 & 47.5 & 5.5 & 18.9 & 2.4 & 43.4 & 6.2 \\
\midrule
\multirow{8}{*}{\textbf{ALBERT}} 
& \textbf{PivotAttack} & \textbf{11.7} & \textbf{1.0} & \textbf{41.0} & \textbf{1.7} & \textbf{39.4} & 6.1 & \textbf{15.4} & \textbf{1.3} & \textbf{51.1} & 6.2 \\
& LimeAttack & 11.4 & 2.4 & 40.3 & 2.5 & 38.6 & \textbf{5.1} & 15.0 & 2.0 & 47.0 & 5.9 \\
& TextHacker & 11.0 & 2.5 & 38.3 & 3.2 & 30.5 & 7.0 & 14.5 & 2.5 & 35.5 & 7.0 \\
& LeapAttack & 11.1 & 2.0 & 37.3 & 3.4 & 35.2 & 5.6 & 15.1 & 2.4 & 42.8 & 6.4 \\
& TextHoaxer & 11.0 & 2.2 & 34.5 & 3.4 & 30.1 & 5.3 & 14.7 & 2.3 & 34.8 & 6.5 \\
& HLBB & 11.5 & 2.8 & 34.2 & 3.8 & 30.1 & 6.0 & 15.3 & 2.7 & 34.1 & 6.5 \\
& VIWHard & 11.5 & 1.5 & 40.5 & 1.7 & 32.0 & 5.2 & 14.9 & 1.4 & 45.2 & 6.3 \\
& HyGloadAttack & 11.6 & 1.8 & 36.9 & 3.2 & 36.6 & 5.5 & 15.2 & 2.3 & 38.1 & 6.6 \\
\bottomrule
\end{tabular}
}
\caption{Comprehensive performance comparison (ASR \% and Pert \%) across all victim models and datasets under a 100-Query Budget. }
\label{table:appendix_results}
\end{table*}

\section{Evaluation Metrics} 
\label{appendix:evaluation-metric}
Consistent with prior work such as LimeAttack and Leapattack, we employ Attack Success Rate (ASR), Perturbation Rate (Pert), and Semantic Similarity (Sim) as evaluation metrics. ASR measures the effectiveness of the attack by calculating the proportion of successfully misclassified examples. To assess the quality and stealthiness of the adversarial text, we use Pert to quantify the magnitude of modifications (i.e., the fraction of tokens changed) and Sim to evaluate the degree of semantic preservation relative to the original input.

\section{Implementation Details} 
\label{appendix:implementation-detail}
We set the query budget to $B = 100$, the KL-LUCB quota parameter to $\gamma = 0.8$, the retention precision threshold to $\tau = 0.85$, the maximum confidence error to $\epsilon = 0.9$, the tolerable prediction error to $\delta = 0.85$, the sampling count to $N = 5$, the substitution candidate set size to $M = 50$, and the base perturbation rate to $\rho = 0.1$. All experiments were conducted on an NVIDIA RTX 3080 Ti GPU (13 GB), and the results are averaged over three independent runs.

\section{Additional Experimental Results on WordCNN and ALBERT}
\label{app:more_result}
To verify generalization, we evaluate {\OURS} on WordCNN and ALBERT (Table~\ref{table:appendix_results}). Consistent with main results, {\OURS} maintains dominance across diverse architectures. On WordCNN (Yelp), {\OURS} matches the top baseline's ASR (12.9\%) but with nearly half the perturbation (1.6\% vs. 3.0\%). On ALBERT (SST-2), {\OURS} leads with 51.1\% ASR compared to the runner-up's 47.0\% (LimeAttack). These results confirm that {\OURS}'s superiority is model-agnostic, effective against both convolutional networks and lightweight Transformers.

\section{Adversary Quality} 
\label{appendix:adversary-quality}
To balance effectiveness and stealth, adversarial samples must preserve semantic similarity with the original text. We further evaluate attacks on the Yelp dataset with BERT as the victim model, using additional metrics: semantic similarity (Sim.) via USE~\citep{bowman2015large} and grammatical error rate (Gram.) via LanguageTool\footnote{\href{https://languagetool.org}{https://languagetool.org}}. As shown in Table \ref{tab:qality}, {\OURS} maintains superior performance across these metrics while achieving the highest attack success.

\begin{table}[ht]
\begin{adjustbox}{max width=\linewidth}
\centering
\small
\begin{tabular}{lcccc}
\toprule
\textbf{Attack} & \textbf{ASR.\,$\uparrow$} & \textbf{Pert.\,$\downarrow$} & \textbf{Sim.\,$\uparrow$} & \textbf{Gram.\,$\downarrow$} \\
\midrule
\textbf{{\OURS}} & \textbf{9.7} & \textbf{1} & \textbf{99.5} & 0.31 \\
LimeAttack  & 7.2 & 2.6 & 99.3           & 0.34 \\
TextHacker  & 7.4 & 2.5 & 99.4           & \textbf{0.27} \\
LeapAttack  & 6.1 & 2.6 & 99.3          & 0.55 \\
TextHoaxer  & 5.8 & 3.4 & 99.4  & 0.73\\
HLBB        & 6.4 & 2 & 99.4           & 0.41 \\
\bottomrule
\end{tabular}
\end{adjustbox}
\caption{Adversarial Sample Quality Comparison on BERT (Yelp, 100-Query Budget)}
\label{tab:qality}
\end{table}

\section{Query Budget Results}
\label{appendix:query-budget}
\begin{figure}[ht]
  \centering
  \begin{subfigure}{0.49\linewidth}
    \includegraphics[width=\linewidth]{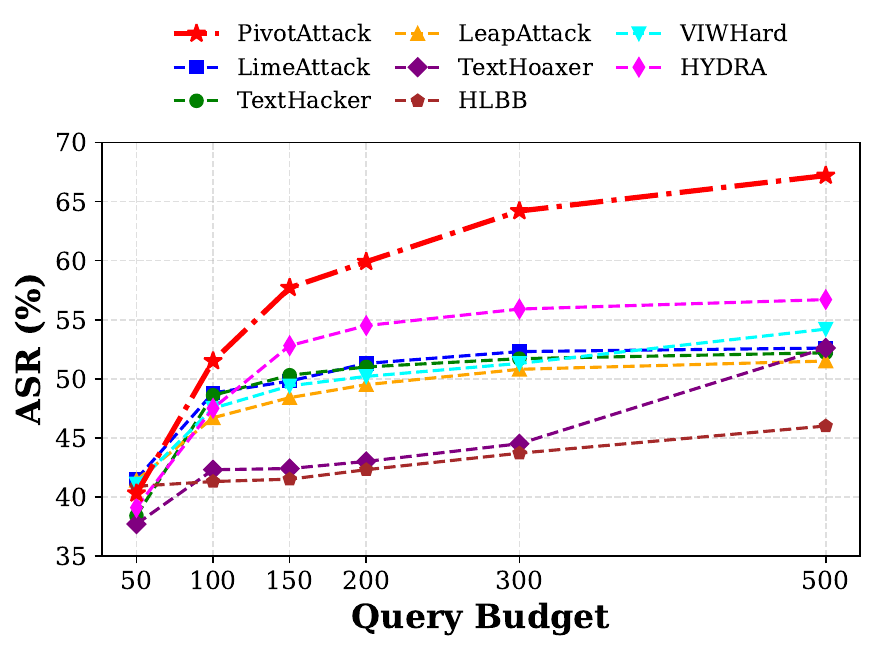}
    \caption{MR (CNN)}
  \end{subfigure}\hfill
  \begin{subfigure}{0.49\linewidth}
    \includegraphics[width=\linewidth]{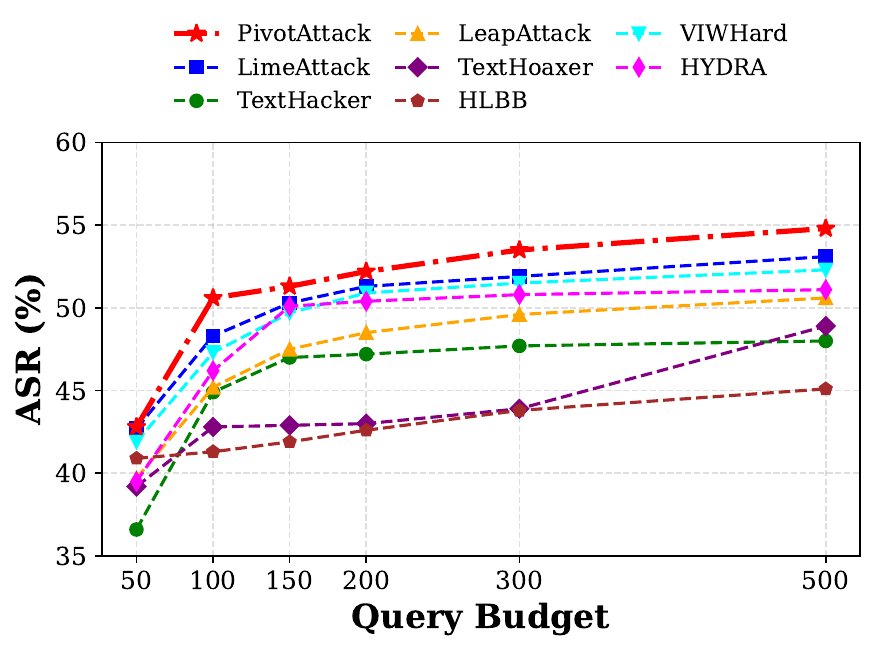}
    \caption{MR (LSTM)}
  \end{subfigure}\hfill
  \begin{subfigure}{0.49\linewidth}
    \includegraphics[width=\linewidth]{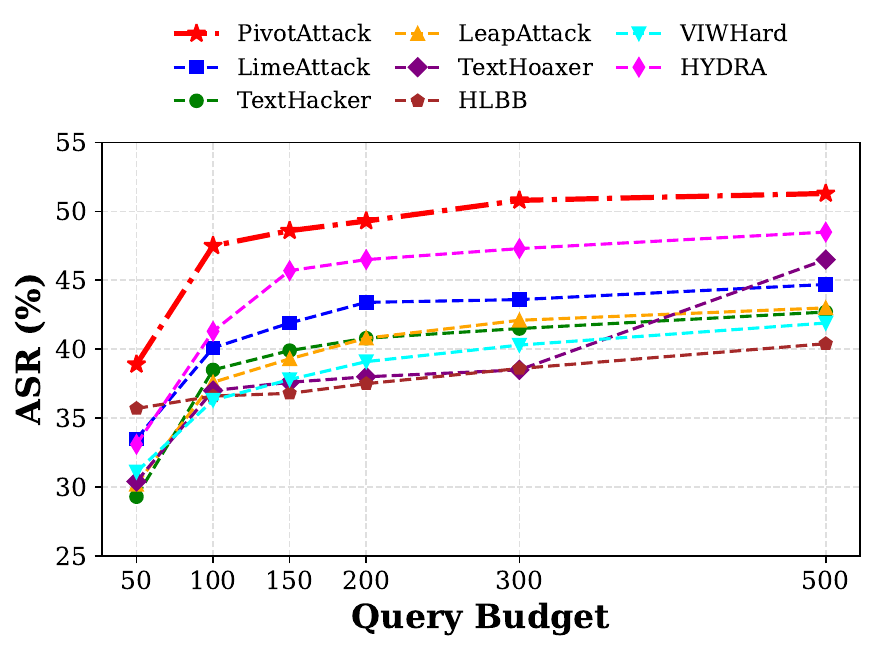}
    \caption{MR (BERT)}
  \end{subfigure}\hfill
  \begin{subfigure}{0.49\linewidth}
    \includegraphics[width=\linewidth]{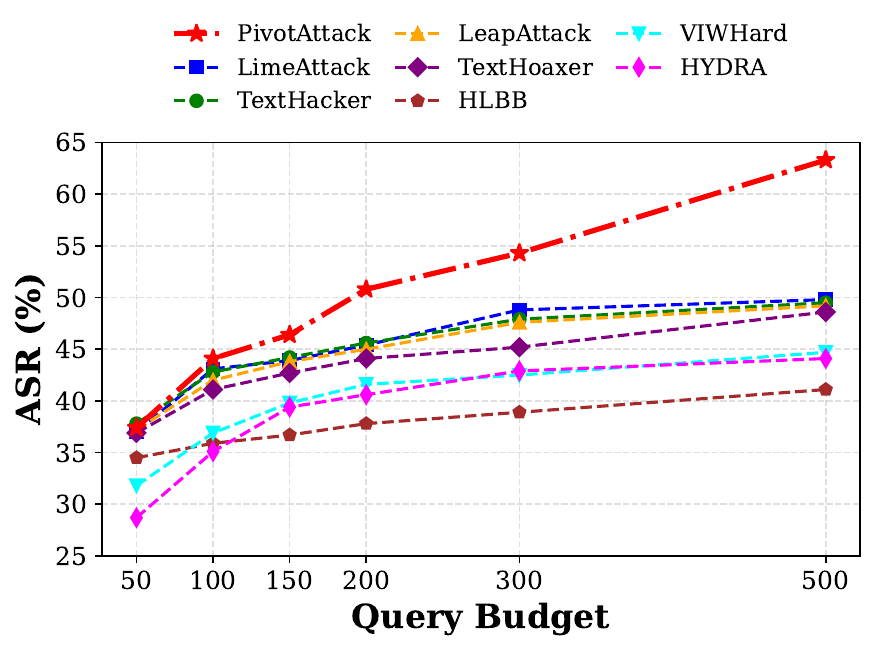}
    \caption{MR (Qwen2.5-ZS)}
  \end{subfigure}
   \begin{subfigure}{0.49\linewidth}
    \includegraphics[width=\linewidth]{figures/llmft-mr.pdf}
    \caption{MR (Qwen2.5-FT)}
  \end{subfigure}
  \begin{subfigure}{0.49\linewidth}
    \includegraphics[width=\linewidth]{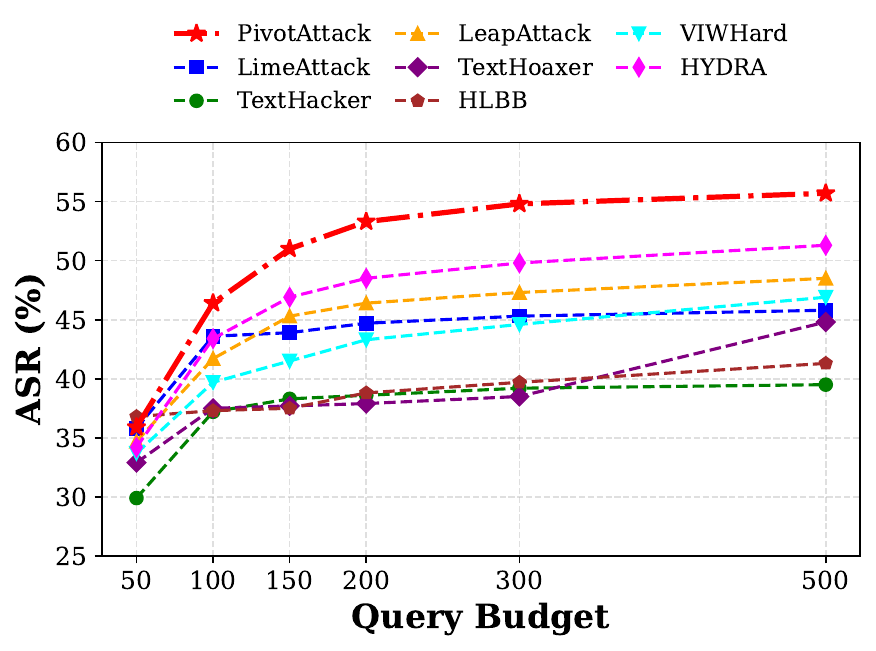}
    \caption{SST-2 (CNN)}
  \end{subfigure}\hfill
  \begin{subfigure}{0.49\linewidth}
    \includegraphics[width=\linewidth]{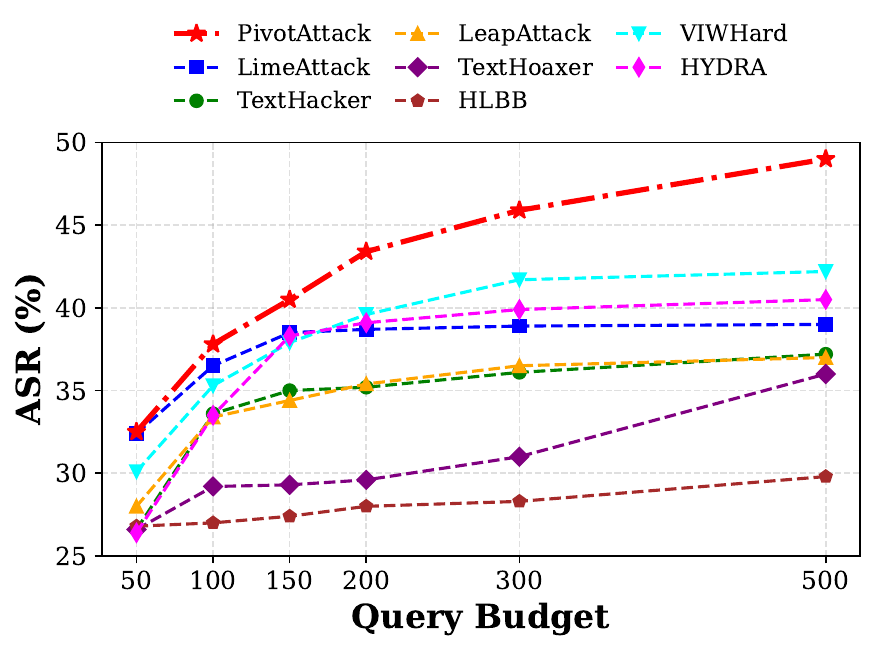}
    \caption{SST-2 (LSTM)}
  \end{subfigure}\hfill
  \begin{subfigure}{0.49\linewidth}
    \includegraphics[width=\linewidth]{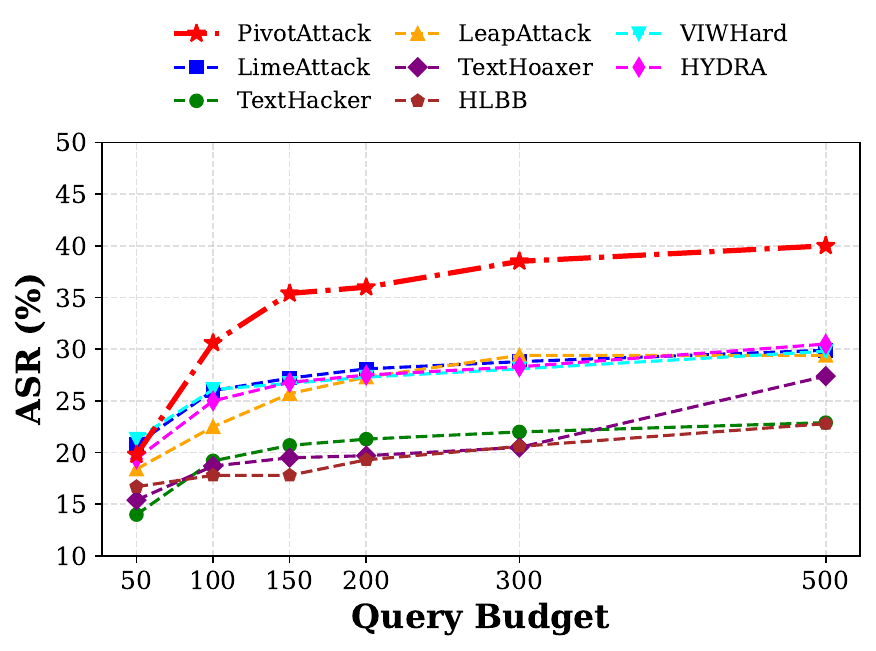}
    \caption{SST-2 (BERT)}
  \end{subfigure}\hfill
  \begin{subfigure}{0.49\linewidth}
    \includegraphics[width=\linewidth]{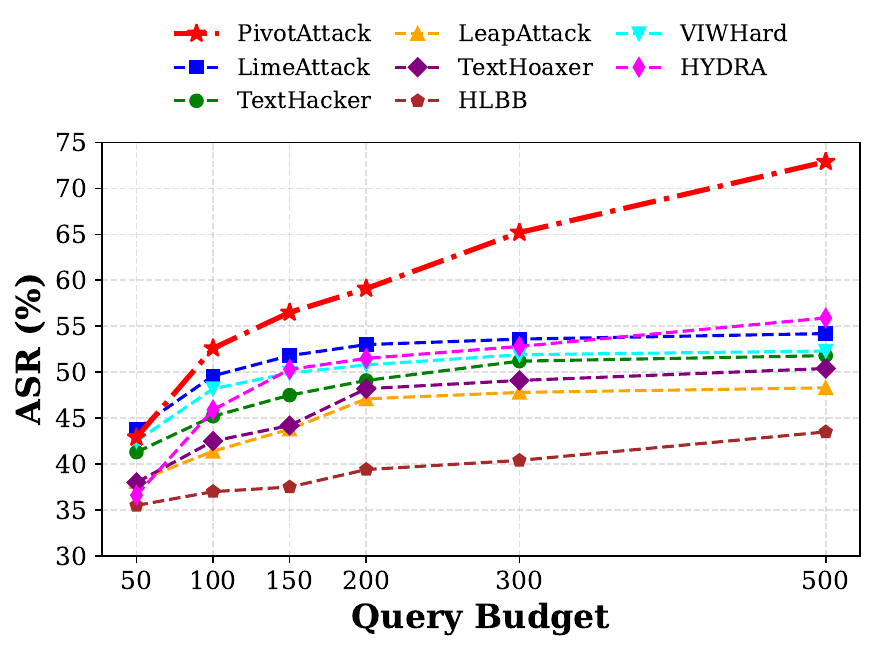}
    \caption{SST-2 (Qwen2.5-ZS)}
  \end{subfigure}
    \begin{subfigure}{0.49\linewidth}
    \includegraphics[width=\linewidth]{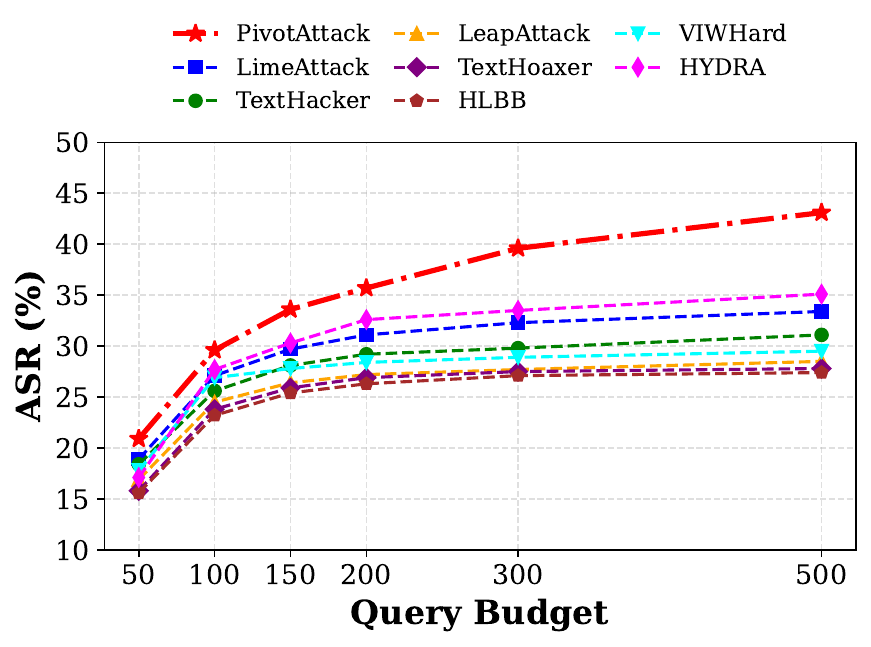}
    \caption{SST-2 (Qwen2.5-FT)}
  \end{subfigure}
  \caption{ASR of different models under different query budgets on MR and SST-2.}
  \label{fig:asr-grid-2x4}
\end{figure}

Additional results on the MR and SST-2 datasets under varying query budgets (i.e., (B = 50, 100, 150, 200, 300, 500)) are presented in Figure~\ref{fig:asr-grid-2x4}. As shown, {\OURS} consistently surpasses all baseline methods across different datasets and victim models.

\section{Human Evaluation Details}
\label{app:human}
\begin{figure}[ht]
\centering
\begin{subfigure}[t]{0.48\textwidth}
  \begin{tcolorbox}[title= {it 's hard to resist his enthusiasm , even if the filmmakers come up with nothing original in the way of slapstick sequences},
                     width=\linewidth]
  \footnotesize\ttfamily\raggedright
  Method1 selection: resist, hard
  \end{tcolorbox}

  \begin{tcolorbox}[title= {greatly impressed by the skill of the actors involved in the enterprise},
                     width=\linewidth]
  A  greatly\\  B impressed\\  C  skill\\  D  actors\\  E  enterprise\\
  \footnotesize\ttfamily\raggedright
  (Answer: B, C)
  \end{tcolorbox}
  \caption{Example of {\OURS}}
  \label{fig:human-anchor}
\end{subfigure}\hfill
\begin{subfigure}[t]{0.48\textwidth}
  \begin{tcolorbox}[title= {it 's hard to resist his enthusiasm , even if the filmmakers come up with nothing original in the way of slapstick sequences},
width=\linewidth]
  \footnotesize\ttfamily\raggedright
  Method2 selection: even, it
  \end{tcolorbox}

  \begin{tcolorbox}[title= {greatly impressed by the skill of the actors involved in the enterprise},
                     width=\linewidth]
  A  greatly\\  B impressed\\  C  skill\\  D  actors\\  E  enterprise\\
 \footnotesize\ttfamily\raggedright
  (Answer: A, D)
  \end{tcolorbox}
  \caption{Example of LimeAttack}
  \label{fig:human-lime}
\end{subfigure}

\caption{Example of the Human Evaluation Survey}
\label{fig:human-combined}
\end{figure}
 
To assess the interpretability of {\OURS}, we conducted a survey with 10 computer science students, comparing it against LimeAttack. The algorithm names were masked. As shown in Figure~\ref{fig:human-anchor}, participants were first asked to review 10 prediction examples, each consisting of a sentence along with the two most important keywords selected by the algorithm. Subsequently, in the testing phase, participants were required to complete 20 multiple-choice questions. Each question provided five candidate keywords, from which participants had to select the two words they believed the algorithm would identify as most important. In this experiment, selecting at least one correct keyword was considered a correct response.

\end{document}